\documentclass[accepted]{uai2023} 

\usepackage[american]{babel}

\usepackage{natbib} 
\usepackage{booktabs} 
\usepackage{tikz} 
\usepackage{hyperref}
\usepackage{url}
\usepackage{color}
\usepackage{tcolorbox}
\usepackage{CJK}
\usepackage{adjustbox}
\usepackage{algorithm}
\usepackage{algorithmic}
\usepackage{graphicx}
\usepackage{booktabs}
\usepackage{threeparttable}
\usepackage{siunitx}
\usepackage{lipsum}
\usepackage{multirow}
\usepackage{bm}
\usepackage{amssymb}
\usepackage{amsmath,amsthm,mathtools}
\usepackage[normalem]{ulem}
\usepackage{commath}
\usepackage{wrapfig,lipsum}
\usepackage{enumitem}
\usepackage{subfig}

\theoremstyle{plain}
\newtheorem{theorem}{Theorem}[section]

\theoremstyle{definition}
\newtheorem{definition}[theorem]{Definition}
\newtheorem{assumption}[theorem]{Assumption}
\theoremstyle{remark}

\title{\texttt{CUE}: An Uncertainty Interpretation Framework for Text Classifiers Built on Pre-Trained Language Models}

\author[1]{\href{mailto:<jiazheng.li@kcl.ac.uk>?Subject=UAI 2023 paper}{Jiazheng Li}}
\author[2]{Zhaoyue Sun}
\author[3]{Bin Liang}
\author[1]{Lin Gui}
\author[1,2,4]{Yulan He}
\affil[1]{%
 Department of Informatics, King’s College London, UK
}
\affil[2]{%
    Department of Computer Science, University of Warwick, UK
}
\affil[3]{%
   Joint Lab of HITSZ-CMS, Harbin Institute of Technology, Shenzhen, China
  } 
\affil[4]{%
    The Alan Turing Institute, UK
  }

\begin{document}
\maketitle

\begin{abstract}
Text classifiers built on Pre-trained Language Models (PLMs) have achieved remarkable progress in various tasks including sentiment analysis, natural language inference, and question-answering. However, the occurrence of uncertain predictions by these classifiers poses a challenge to their reliability when deployed in practical applications. Much effort has been devoted to designing various probes in order to understand what PLMs capture. But few studies have delved into factors influencing PLM-based classifiers' predictive uncertainty. In this paper, we propose a novel framework, called \texttt{CUE}, which aims to interpret uncertainties inherent in the predictions of PLM-based models. In particular, we first map PLM-encoded representations to a latent space via a variational auto-encoder. We then generate text representations by perturbing the latent space which causes fluctuation in predictive uncertainty. By comparing the difference in predictive uncertainty between the perturbed and the original text representations, we are able to identify the latent dimensions responsible for uncertainty and subsequently trace back to the input features that contribute to such uncertainty. Our extensive experiments on four benchmark datasets encompassing linguistic acceptability classification, emotion classification, and natural language inference show the feasibility of our proposed framework. Our source code is available at \url{https://github.com/lijiazheng99/CUE}.
\end{abstract}

\section{Introduction}

Text classifiers built on Pre-trained Language Models (PLMs) have made remarkable progress on various Natural Language Processing (NLP) tasks \citep{bert,roberta,albert,distilbert}. However, their deployment in practical applications still faces significant challenges. Of particular concern, these models 
tend to make over-confident predictions in uncertain cases \citep{softmax_uncertain, he-etal-2020-towards,predictive_uncertainty}. Since PLMs have been widely used in various applications, such issues cause concerns about model trustworthiness and transparency, which becomes a barrier to deploying PLMs in sensitive domains such as medicine and finance.  
\begin{figure*}[!t]
  \centering
  \includegraphics[width=\linewidth]{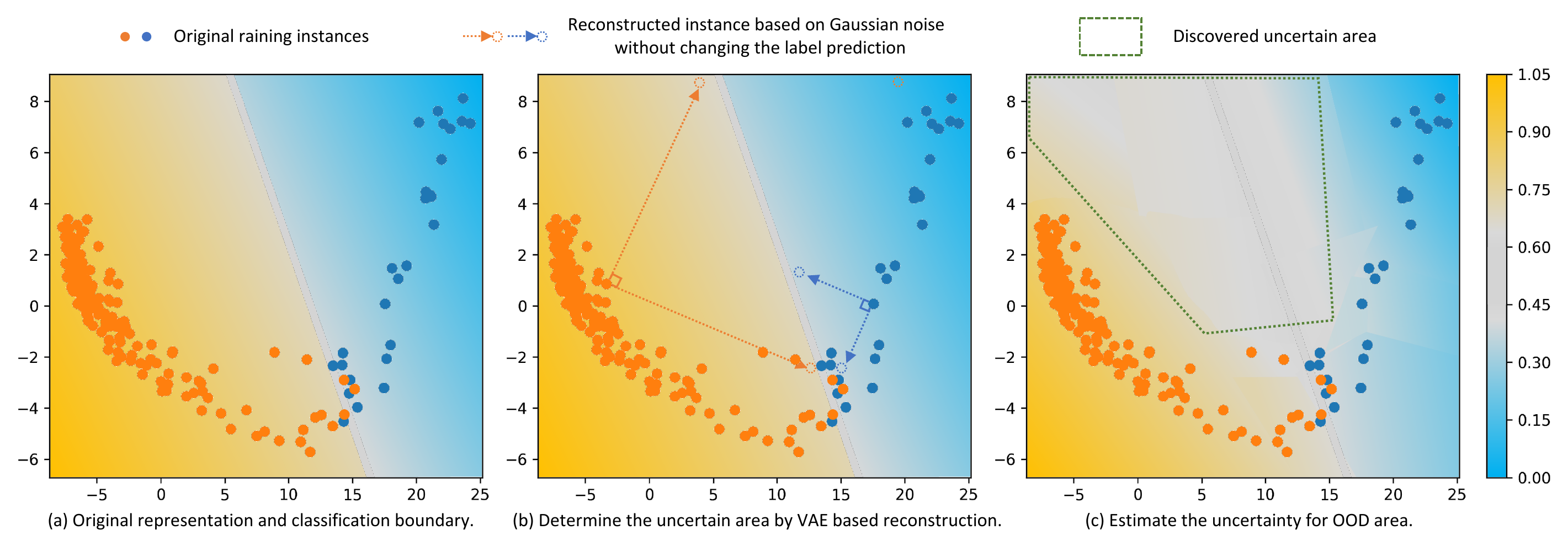}
  \caption{The illustration of proposed framework: \textbf{(a)} The original representations learned from a Pre-trained Language Model (PLM) and the decision boundary separating two classes; \textbf{(b)} We can freeze the PLM-classifier parameters and perturb the PLM-encoded representations in the latent space to increase the aleatoric (data) uncertainty along some isotropic directions while preserving the predictive labels. This would help determine the uncertain areas; \textbf{(c)} Using the representations reconstructed from the perturbed latent space, the uncertainty for out-of-distribution (OOD) areas can be estimated. }
  \label{fig:exampleIllustration}
\end{figure*} 

Predictive uncertainty is generally believed to include two aspects - \emph{aleatoric uncertainty} and \emph{epistemic uncertainty}, where the aleatoric uncertainty measures the data uncertainty due to inherent random effects and is irreducible, while the 
epistemic uncertainty measures the uncertainty caused by the lack of knowledge from data and is reducible \citep{predictive_uncertainty}.
Numerous approaches have been proposed to estimate the predictive uncertainty of deep neural models, such as Deep Ensemble models \citep{deep_ensemble}, Bayesian Neural Networks (BNN) \citep{bnn_wight_uncertainty} and Monte-Carlo (MC) Dropout \citep{monte_carlo_dropout}. 
Similar idea has been applied to PLMs in recent years, to study the uncertainty of text classifiers.
Particularly, quantifying uncertainty in PLM-based classifiers can be done by incorporating weight uncertainty into the PLM architecture. However, uncertainty can only be induced to a certain number of layers (e.g., the last layer of the PLM feature extractor and/or the classification layer) due to a large number of PLM layers and parameters. Alternatively, one could use deep ensembles by aggregating classification results generated from multiple PLM classifiers trained with different initialisation \citep{deep_ensemble}, or apply MC dropout in the inference stage to estimate the uncertainty of PLMs \citep{acl2022_uncertainty_transformers}.

Previous studies \citep{vulic-etal-2020-probing, clark-etal-2019-bert, yang-etal-2021-exploring} have also been devoted to designing various probes in order to understand what PLMs capture. Nevertheless, they largely ignore the interpretation of the source of the uncertainty, i.e., identifying the input features which cause classification uncertainty, which can be crucial for understanding the model and taking appropriate mitigating strategies. In text classification, recent research tried to identify word tokens that lead to uncertainty via perturbations on input sequences \citep{token_level_uncertainty,token_level_classification_interpret}. However, due to the discreteness of textual data, token replacement or removal would require a large search space on the input sequence and incur expensive computational costs. 


In this paper, we aim to interpret the predictive uncertainty on PLM text classifiers 
by identifying the input tokens that cause the uncertainty.
We propose a novel PLM \textbf{C}lassifier \textbf{U}ncertainty \textbf{E}xplanation (\texttt{CUE}) framework built on Variational Auto-encoder (VAE) \citep{vae,scholar_vae} that generates perturbations on latent text representations to induce uncertain predictions.
As shown in Figure \ref{fig:exampleIllustration}, we can perturb the PLM-encoded representations in the latent space to increase the aleatoric uncertainty (data uncertainty) along some isotropic directions while preserving the predictive labels. As will be shown in \textsection{\ref{sec:uncertain_estimation}}, this is equivalent to decreasing the predictive epistemic uncertainty. By examining the difference between the original and the perturbed text representations, a subset of input features (i.e., word tokens) can be identified as the interpretation of the original model’s predictive uncertainty. 
We compared our framework with existing approaches addressing the predictive uncertainty problem on three classification tasks across four benchmark datasets. Extensive experimental results show that our proposed method can identify the source of epistemic uncertainty and calibrate text representations from four commonly used PLMs.  

In summary, our contributions are:
  \textbf{(1)} We propose a novel framework \texttt{CUE} to induce perturbations on PLM-encoded representations for uncertainty interpretation of the PLM-based text classifiers.
  \textbf{(2)} We propose an uncertainty feature identification algorithm to identify token-level features which lead to model predictive uncertainty.
  \textbf{(3)} We validate the effectiveness of our proposed framework by conducting extensive experiments using various classifiers built on four commonly-used PLMs on three different tasks and four datasets with class numbers ranging from 2 to 27. The results show that our proposed framework achieves lower expected calibration errors compared to existing approaches such as label smoothing, MC dropout, and BNN. 
To the best of our knowledge, our framework is the first to study the token interpretation of PLM-based classifiers' predictive uncertainty from the representation space, without editing the semantic meaning of the original input text.

\section{Related work}
Our work is related to two lines of research, interpretation of PLMs and uncertainty estimation in ML.
\paragraph{Interpretation of PLMs}
Transformer-based language models have achieved impressive performance across various NLP tasks \citep{bert,roberta}. However, the complex structure of these models has raised concerns about model transparency and reliability. Thus, there has been growing interest in developing methods to interpret PLMs. For example, \citet{clark-etal-2019-bert} proposed an attention-based visualisation method to interpret the model parameters by probing the feature space to determine the potential influence of the model output. \citet{Brunner2020On} studied the identifiability of attention weights in the BERT model and found that the distribution of self-attentions cannot be directly used as an interpretation. There has also been work focusing on interpreting the representations from PLMs \citep{zhou-srikumar-2021-directprobe} and attention weights \citep{sun-marasovic-2021-effective,marecek-rosa-2019-balustrades}.
\paragraph{Uncertainty Estimation}
As the interpretation of PLMs cannot provide prediction confidence directly, much effort has been devoted to developing approaches for uncertainty estimation of neural models. A straightforward approach to uncertainty estimation is 
using deep ensemble models 
\citep{deep_ensemble}. 
Various Bayesian inference methods have also been developed to prevent overfitting  
by attaching distributions to parameters in standard networks and estimating parameters via posterior inference \citep{bnn_wight_uncertainty}. Alternatively, uncertainty estimation can be performed using MC dropout \citep{monte_carlo_dropout}, which performs multiple stochastic forwards passes with dropout in a network during the inference stage to produce an ensemble of predictions. 
Other approaches to uncertainty estimation include prior networks \citep{predictive_uncertainty}. 
Taking advantage of the development of transformer \citep{transformer}, there has been increasing interest in investigating classification uncertainty of language models \citep{quantify_uncertainty_nlp,desai-durrett-2020-calibration}. Various methods have been developed, partly inspired by the research in computer vision, from uncertainty quantification via input marginalization \citep{token_level_uncertainty,token_level_classification_interpret} to MC dropout and Bayesian inference methods such as SNGP \citep{how_certain_transformer,acl2022_uncertainty_transformers,sngp}. 
Nevertheless, the aforementioned approaches cannot identify the cause of the uncertainty. 

To overcome the limitation of existing methods, we propose an uncertainty analysis framework \texttt{CUE} built on VAE \citep{vae,scholar_vae,gaussian_process_uncertainty}, in which noise can be generated by perturbing the latent representation space. This allows us to disentangle the source of uncertainty via text representation dimensions and study PLM-based classifiers' predictive uncertainty at both the sequence- and the token-level.

\section{Background}
\begin{figure*}[!t]
  \centering
  \includegraphics[width=0.84\linewidth]{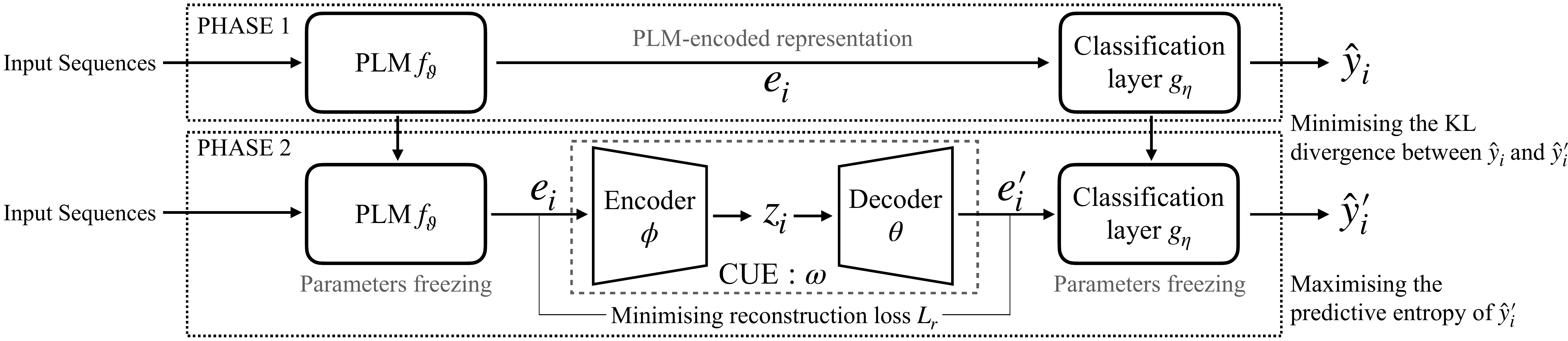}
\caption{\texttt{CUE} requires two phases of training: \textbf{Phase 1}: Fine-tune a classifier built on a PLM on a target dataset; \textbf{Phase 2}: Freeze the PLM and classification layer parameters and plug-in the \texttt{CUE} module to train the model that perturbs the latent representation $\bm{z}_i$ to generate $\bm{e}'_i$ which leads to uncertainty fluctuation.}
\label{fig:pipeline}
\end{figure*} 

\subsection{Problem setup}

We are given a labelled text classification dataset, where $X$ is input text and $Y$ is the label set. $\forall \{x_i, y_i\} \in X \times Y$, where $(x_i,y_i), i = 1,2, ...,N$, is an i.i.d. realisation of the random variables, $P(Y|X) \sim \mathbb{D}$, where $\mathbb{D}$ is the unknown ground truth conditional distribution of class labels. To train a text classifier built on a PLM, we need to find an optimal feature extraction function $f$ and a classification layer $g$ with trainable parameter $\bm{\vartheta}$ and $\bm{\eta}$, respectively: $x_i \stackrel{f_{\bm{\vartheta}} (\cdot) }{\longrightarrow} \bm{e}_i \stackrel{g_{\bm{\eta}(\cdot)}} {\longrightarrow} \hat{y}_i$, which first encodes text into a  representation $\bm{e}_i$ and then outputs a probability distribution over the label set with the predicted output close to the desired true label $y_i$. In this work, we take one step further to analyse the potential uncertainty in the two stages of the learning process: 1) in $\bm{\eta}$: $\bm{e}_i \stackrel{g_{\bm{\eta}(\cdot)}} {\longrightarrow} \hat{y}_i$, which dimension(s) in $\bm{e}_i$ is the source of uncertainty in prediction; and 2) in $\bm{\vartheta}$: $x_i \stackrel{f_{\bm{\vartheta}} (\cdot) }{\longrightarrow} \bm{e}_i$, which input tokens cause the uncertainty. Before we detail our proposed uncertainty estimation approach, we give the formal definition of uncertainty first.
\subsection{Uncertainty Estimation}
\label{sec:uncertain_estimation}
According to established definitions found in prior literature, uncertainty can be defined based on the probability of predictive error \citep{sullivan2015introduction}, the mean squared error (MSE) \citep{mse_uncertainty}, or the conditional entropy \citep{entropy_uncertainty}. 
We adopt the MSE-based definition as a representative measure of uncertainty, which is chosen without compromising the generality of our approach. 
\begin{definition}
$\forall P(y|x) \in \mathbb{D}$, the predictive epistemic uncertainty can be defined by $\mathbb{E}\big[\big(\mathbb{E}[y] - \hat{y}\big)^2\big]$.
\end{definition}
Here, $\hat{y}$ is the class label for input $x$ predicted by the trained classifier. $\mathbb{E}[y]$ is the expectation of the ground truth label distribution, which is however unknown to the learner, making it impossible to calculate the epistemic uncertainty based on predictive variance directly. 
Therefore, we propose to estimate the uncertainty by decomposing the variance based on the observed training data, $\{x_i,y_i\}_{i=1}^N$, 
which yields:
\begin{equation}
\small
    \mathbb{E}[(y_i - \hat{y}_i)^2] = \underbrace{\mathbb{E}[(y_i - \mathbb{E}[y])^2]}_{\rm aleatoric\,uncertainty} + \underbrace{\mathbb{E}[(\mathbb{E}[y] - \hat{y}_i)^2]}_{\rm epistemic\,uncertainty}
    \label{eq:predVariance}
\end{equation}
Since the first term, $\mathbb{E}[(y_i - \mathbb{E}[y])^2]$, contains the observed $y_i$, it can be defined as the aleatoric uncertainty. The detailed derivation of Eq. (\ref{eq:predVariance}) can be found in our Supplementary Material Section 1.1. Similar to the setup in \citep{heiss2023nomu}, if we assume the conditional distribution of class labels follows a Gaussian distribution:
\begin{assumption}
\label{assum:gaussian}
    $\forall P(y_i|x_i) \in \mathbb{D}$, the true label distribution for a give data $x_i$ follows a Gaussian noise based generating process: $P(y_i|x_i) = \bm{\mu}_y + \bm{\varepsilon}_{y_i}$, 
    where $\bm{\mu}_y = \mathbb{E}[y] $ and the noise $\bm{\varepsilon}_{y_i}$ follows a Gaussian distribution of $\mathcal{N}(0,\bm{\sigma}_y^2)$ and $\bm{\sigma}_y^2 = \mathbb{E}[(\mathbb{E}[y] - y_i)^2] $.
\end{assumption}
Then, the epistemic uncertainty can be written as:
\begin{equation}
    \mathbb{E}\big[(\mathbb{E}[y] - \hat{y}_i)^2\big] = \mathbb{E}[(y_i - \hat{y}_i)^2] - \bm{\sigma}_y^2.
    \label{eq:epistemic}
\end{equation}
Here, the term $\mathbb{E}[(y_i - \hat{y}_i)^2]$ in the Eq. (\ref{eq:epistemic}) is the empirical MSE on the training data which can be optimised in the training process. The term $\bm{\sigma}_y^2$ is based on the true label distribution which is unseen to the learner. We need to clarify that $P(y_i|x_i)$ can be larger than $1$ under the assumption. Therefore, there is necessary to stack a normalisation layer before the prediction to guarantee the sum of predictive probabilities for different class labels is $1$. 

We assume that the empirical MSE has been minimised by the trained PLM-based classifier with parameters $\bm{\vartheta}$ and $\bm{\eta}$. To minimise the epistemic uncertainty given by Eq. (\ref{eq:epistemic}) for a given $x_i$ and its corresponding representation $\bm{e}_i$, we have to increase $\bm{\sigma}_y^2$, which however cannot be calculated directly. We propose to use a VAE-based generative model parameterised by $\bm{\omega}$ to reconstruct $\bm{e}_i$ by adding Gaussian noise while preserving the predictive label, resulting in $\bm{e}'_i$. The reconstructed representation should be similar to the original input representation, $\bm{e}'_i\approx \bm{e}_i$, and the predictive class label distribution from $\bm{e}'_i$, $\hat{y}'_i = g_{\bm{\eta}}(\bm{e}'_i)$, should be close to $\hat{y}_i= g_{\bm{\eta}}(\bm{e}_i)$, $\hat{y}'_i \approx \hat{y}_i$. 
This allows us to manipulate the latent code of the VAE to increase the variance of the Gaussian noise, which leads to the resulting label distribution 
closer to a uniform distribution in the out-of-distribution (OOD) area, thus achieving a lower epistemic uncertainty. Accordingly, we define the learning objective function as:

\textbf{Learning objective:} $\forall P(y_i|x_i) \in \mathbb{D}$ under the Assumption \ref{assum:gaussian}, the learning objective is to:
\begin{equation}
\small
     \mathop{min}_{\bm{\omega}} \mathbb{E}[(y_i - \hat{y}_i)^2] - \mathcal{H}_{e'}(\hat{y}),  \quad
    s.t. \quad \bm{e}_i \approx \bm{e}'_i, \quad 
    \hat{y}_i \approx \hat{y}'_i,
\end{equation}
where $\bm{\omega}$ denotes the parameters of VAE, $\bm{e}_i=f_{\bm{\vartheta}}(x_i)$, $\bm{e}'_i=f_{\bm{\omega}}(\bm{e}_i)$, $\hat{y}_i= g_{\bm{\eta}}(\bm{e}_i)$, $\hat{y}'_i = g_{\bm{\eta}}(\bm{e}'_i)$, $\mathcal{H}_{e'}(\hat{y})$ is the estimated entropy by the predictive label distribution from the reconstructed $\bm{e}'_i$, which approximates the variance of the true label distribution. The above learning objective can be formulated using the method of Lagrange multipliers: 
{\small 
\begin{align}
    \mathcal{L}(x_i,y_i) = &- \mathbb{E}[ \mathcal{H}_{e'}(\hat{y}_i)]     + \lambda_1 \mathbb{E}[(f_{\bm{\vartheta}}(x_i)-f_{\bm{\omega}}(f_{\bm{\vartheta}}(x_i)))^2] \notag \\
    &+ \lambda_2 \mbox{KL}\big( \hat{y}'_i || \hat{y}_i \big)
    \label{eq:learningobjective}
\end{align}}
Therefore, by optimising Eq. (\ref{eq:learningobjective}), we can obtain an alternative representation of $\bm{e}'_i$ with the predictive distribution of $\hat{y}'_i$ using the parameters $\bm{\omega}$, where the lower bound of the epistemic uncertainty can be obtained by increasing the aleatoric uncertainty defined by $\mathcal{H}_{e'}(\hat{y}_i)$. 
In the next section, we show how each term in Eq. (\ref{eq:learningobjective}) can be defined in our VAE-based uncertainty interpretation framework \texttt{CUE}.

\section{Uncertainty Interpretation}

In this paper, we are interested in interpreting model uncertainty, that is, what input features lead to the predictive uncertainty. 
To this end, we propose a VAE-based uncertainty interpretation framework \texttt{CUE}, as shown in Figure \ref{fig:pipeline}. Rather than directly perturbing the input features, perturbations can be done in the latent space in \texttt{CUE} to generate the modified input representation such that it still resides on the original data manifold while the model's predictive epistemic uncertainty on the modified input is reduced. By examining the difference between the original and the perturbed text representations, a subset of input features (i.e., word tokens) can be identified as the interpretation of the original model's predictive uncertainty. 

We will first present how to generate perturbations on latent space in order to cause the prediction uncertainty change (\textsection{\ref{sec:counterfactual}}). We will then describe how to identify input features that lead to original prediction uncertainty to facilitate the interpretation of model predictive uncertainty  (\textsection{\ref{sec:inputFeatureIdentification}}).

\subsection{Latent Space Perturbation for Epistemic Uncertainty Reduction}\label{sec:counterfactual}
Once a classifier built on a PLM is fine-tuned on a target dataset, we freeze the parameters of the PLM and the classification layer and then insert the \texttt{CUE} between the PLM last layer and the task-specific classification layer. 
The PLM-encoded representation $\bm{e}_i$ is mapped to a latent vector, denoted by $\bm{z}_i$, via \texttt{CUE} which consists of two networks. 

\textbf{The \emph{encoder} network $\bm{\phi}$}, learns the distribution of a lower dimensional latent variable $\bm{z}_i$ given the PLM-encoded representation by a random Gaussian noise $\epsilon$: $\bm{z}_i=\bm{\mu_{\phi}}(\bm{e_i})+\epsilon\cdot\bm{\sigma_{\phi}}(\bm{e_i})$, i.e., $\bm{z}_i \sim \mathcal{N}(\bm{\mu_{\phi}},\bm{\sigma_{\phi}}^2)$. 

\textbf{The \emph{decoder} network $\bm{\theta}$}, reconstructs the text representation given the latent variable $\bm{z}_i$, defined as $\bm{e}_i' = p_{\bm{\theta}}(\bm{z}_i)$. Although $p_{\bm{\theta}}(\bm{z}_i)$ can be any decoding network, our implementation utilises a linear mapping $\bm{W_{\theta}}$ without a bias term. The benefit is that $\bm{W_{\theta}}$ can be treated as a set of learnable vectors and the reconstructed text representation $\bm{e}_i'$ can be written as a linear combination of the decoded output generated from each of the latent dimensions of $\bm{z}_i$. As will be discussed in \textsection{\ref{sec:inputFeatureIdentification}}, such a decomposition form of decoding as illustrated in Eq. (\ref{eq:GMPR}) allows the identification of latent dimensions of $\bm{z}_i$ which causes predictive uncertainty. 
The VAE parameters are denoted as $\bm{\omega} = \{\bm{\phi}, \bm{\theta}\}$.  The classifier's prediction on the reconstructed representation $\bm{e}'_i$ is denoted as $\hat{y}'_i = g_{\bm{\eta}}(\bm{e}'_i)$. Here, we choose to use the Softmax based prediction layer to normalise the predictive probability, but the representation $\bm{e}'_i$ before the normalisation should follow the Gaussian distribution since it is captured by a linear combination of Gaussians. 
Besides, the latent representation $\bm{z}_i$ can be perturbed which leads to uncertain predictions bounded by a uniform distribution probability, $\log K$ ($K$ is the label set size)\footnote{The proof is shown in Supplementary Material Section 1.2.}. 
For the training of the \texttt{CUE} model, we define various loss terms in Eq. (\ref{eq:learningobjective}) below:

\textbf{Minimum change on both the perturbed representation and the model prediction}. The reconstructed $\bm{e}'_i$ should be similar to the original $\bm{e}_i$.
\begin{equation}
\small
    \mathcal{L}_r = \norm{\bm{e}'_i-\bm{e}_i}^2 \label{eq:reconstruction}
\end{equation}
The prediction, $p(\hat{y}'_i|\bm{e}'_i)$, based on the reconstructed $e'_i$, should be close to the origin prediction $p(\hat{y}_i|\bm{e}_i)$.
\begin{equation}
\small
\mbox{KL}(\hat{y}'_i || \hat{y}_i) = \sum_{k=1}^K p(\hat{y}'_{i_k}|\bm{e}'_i)\log \frac{p(\hat{y}'_{i_k}|\bm{e}'_i)}{p(\hat{y}_{i_k}|\bm{e}_i) },
\end{equation}
where $K$ denotes the size of the class label set. 

\textbf{Predictive Entropy Increment}.
We need to increase the predictive entropy $\mathcal{H}_{e'}(\hat{y}'_i)$ calculated based on the reconstructed input representation $\bm{e}'_i$, which approximates the variance of the true label distribution, $\bm{\sigma}_y^2$, in order to decrease the model epistemic uncertainty defined in Eq. (\ref{eq:epistemic}). 
\begin{equation}
\small
    \mathcal{H}_{e'}(\hat{y}'_i) = -\sum_{k=1}^K p(\hat{y}'_{i_k}|\bm{e}'_i)\log p(\hat{y}'_{i_k}|\bm{e}'_i)
\end{equation}
In addition, 
we incorporate an orthogonality constraint within the decoder to encourage independence among dimensions of the latent variable: 
\begin{equation}
\small
    \mathcal{L}_o = \|{\rm \bf I} - \bm{W_{\theta}}\times \bm{W_{\theta}}^{\mathsf{T}}\|\label{eq:orthogonality}
\end{equation}
\noindent where {$\rm \bf I$} is an identity matrix, $W_{\theta}$ is the weights in the decoder. 
The final objective function is then defined as:
\begin{equation}
\small
     \mathcal{L} = \gamma_1\mathcal{L}_r + \gamma_2\mbox{KL}(\hat{y}'_i || \hat{y}_i ) - \gamma_3\mathcal{H}_{e'}(\hat{y}'_i) + \gamma_4 \mathcal{L}_o
     \label{eq:finalObjective}
\end{equation}

where the $\gamma$ coefficients are used to balance various loss terms. Minimising the loss function defined in Eq. (\ref{eq:finalObjective}) is equivalent to introducing perturbation in the latent space so as to increase the predictive entropy. We can use the reconstruction error $||\bm{e}'_i-\bm{e}_i||^2$ to represent the perturbed noise that leads to predictive uncertainty difference $\Delta \mathcal{H}$. As will be shown in Supplementary Material Section 1.2, $\Delta\mathcal{H}$ is proportional to the reconstruction error $||\bm{e}'_i-\bm{e}_i||^2$. As such, the reconstruction error can be used to interpret the predictive uncertainty. 
By retracing alterations made in the input feature space, we can effectively identify features which cause the uncertainty. To the best of our knowledge, we are the first to apply perturbations in the latent representation space to interpret the predictive uncertainty associated with PLM-based classifiers.
\subsection{Input Feature Identification for Uncertainty Interpretation}
\label{sec:inputFeatureIdentification}

\begin{table*}[!ht]
\centering
\resizebox{\linewidth}{!}{
\begin{tabular}{lcccccccc}
\toprule
                 & \multicolumn{4}{c}{CoLA}                                                              & \multicolumn{4}{c}{GoEmotions}                                                                       \\
\cmidrule(lr){2-5} \cmidrule(lr){6-9}  
Model             &  Acc    &  F1  & $\mathcal{H}$  & ECE$\downarrow$    &  Acc    &  F1  & $\mathcal{H}$  & ECE$\downarrow$   \\ \midrule
\midrule
ALBERT (11M)          &0.7923\small{±0.0192}&0.8624\small{±0.0106}&0.4650\small{±0.0695}&0.0834\small{±0.0179} & 0.6193\small{±0.0051}&0.4545\small{±0.0106}&0.3574\small{±0.0042}&0.0446\small{±0.0080}   \\  
ALBERT Label Smoothing&0.7699\small{±0.0452}&0.8508\small{±0.0195}&0.5712\small{±0.1834}&0.0625\small{±0.0345} & 0.6219\small{±0.0025}&0.4579±\small{0.0133}&0.3651\small{±0.0163}&0.0353\small{±0.0146} \\  
ALBERT MC Dropout     &0.7893\small{±0.0109}&0.8583\small{±0.0065}&0.4575\small{±0.0589}&0.0854\small{±0.0223} & 0.6152\small{±0.0055}&0.4448\small{±0.0144}&0.3697\small{±0.0188}&\textbf{0.0345}\small{±0.0076} \\   
ALBERT w/ BNN         &0.7973\small{±0.0011}&0.8647\small{±0.0006}&0.4180\small{±0.0015}&0.0936\small{±0.0016} & 0.6187\small{±0.0009}&0.4396\small{±0.0015} &0.3170\small{±0.0003}&0.0864\small{±0.0011}  \\  
ALBERT w/ \texttt{CUE} (Ours) &0.8038\small{±0.0005}&0.8668\small{±0.0004}&\underline{0.5771}\small{±0.0004}&\textbf{0.0444}\small{±0.0031} & 0.6176\small{±0.0021}&0.4567\small{±0.0046}&\underline{0.3814}\small{±0.0294}&0.0395\small{±0.0098}  \\   \midrule
DistilBERT (66M)          &0.7634\small{±0.0032}&0.8479\small{±0.0019}&0.5412\small{±0.0151}&0.0842\small{±0.0060}&0.6231\small{±0.0018}&0.4637\small{±0.0047}&0.3312\small{±0.0024}&0.0566\small{±0.0039}\\   
DistilBERT Label Smoothing&0.7632\small{±0.0033}&0.8477\small{±0.0018}&0.5620\small{±0.0154}&0.0765\small{±0.0063}&0.6233\small{±0.0021}&0.4643\small{±0.0042}&0.3412\small{±0.0033}&0.0520\small{±0.0035} \\  
DistilBERT MC Dropout     &0.7787\small{±0.0241}&0.8559\small{±0.0125}&0.4773\small{±0.0864}&0.0897\small{±0.0110}&0.6224\small{±0.0023}&0.4670\small{±0.0054}&0.3246\small{±0.0033}&0.0623\small{±0.0025} \\  
DistilBERT w/ BNN         &0.7659\small{±0.0016}&0.8491\small{±0.0010}&0.5133\small{±0.0002}&0.0966\small{±0.0021}&0.6237\small{±0.0009}&0.4550\small{±0.0016}&0.3080\small{±0.0002}&0.0802\small{±0.0005} \\ 
DistilBERT w/ \texttt{CUE} (Ours) &0.7831\small{±0.0012}&0.8540\small{±0.0010}&\underline{0.8362}\small{±0.0008}&\textbf{0.0738}\small{±0.0029}&0.6253\small{±0.0017}&0.4517\small{±0.0022}&\underline{0.4457}\small{±0.0004}&\textbf{0.0208}\small{±0.0031}   \\ \midrule
BERT (110M)         & 0.8000\small{±0.0072}	&0.8696\small{±0.0043}&0.4026\small{±0.0317}&0.0995\small{±0.0057} &0.6266\small{±0.0032} &	0.4829\small{±0.0075} &0.3342\small{±0.0049} &	0.0537\small{±0.0047}\\ 
BERT Label Smoothing& 0.8036\small{±0.0085}	&0.8717\small{±0.0044}&0.4099\small{±0.0843}&0.0967\small{±0.0185} &0.6268\small{±0.0040}&	0.4819\small{±0.0086} &0.3444\small{±0.0046} &	0.0492\small{±0.0020}\\ 
BERT MC Dropout     & 0.8008\small{±0.0054}	&0.8703\small{±0.0037}&0.4023\small{±0.0321}&0.0987\small{±0.0100} &0.6266\small{±0.0026} &	0.4889\small{±0.0085} &0.3337\small{±0.0056} &	0.0548\small{±0.0053}\\ 
BERT w/ BNN         & 0.6104\small{±0.1689}	&0.6545\small{±0.3509}&\underline{0.8837}\small{±0.0732}&0.1095\small{±0.1391} &0.6296\small{±0.0008} &	0.4855\small{±0.0008}         &0.3102\small{±0.0001} &	0.0775\small{±0.0014}\\ 
BERT w/ \texttt{CUE} (Ours)        & 0.8123\small{±0.0012}	&0.8762\small{±0.0007}&0.4991\small{±0.0032}&\textbf{0.0677}\small{±0.0056}&0.6282\small{±0.0029} &	0.4712\small{±0.0087} &\underline{0.4433}\small{±0.0159} &\textbf{0.0326}\small{±0.0013}\\ \midrule
RoBERTa (125M)         &0.8050\small{±0.0142}&0.8721\small{±0.0072} &0.3310\small{±0.0472}&0.1100\small{±0.0190}& 0.6226\small{±0.0051}&0.4877\small{±0.0095}   &0.3310\small{±0.0472}&0.0602\small{±0.0073}\\
RoBERTa Label Smoothing&0.8165\small{±0.0128}&0.8788\small{±0.0058}&0.3091\small{±0.0415}&0.1079\small{±0.0120}& 0.6215\small{±0.0021}&0.4866\small{±0.0098}&0.3091\small{±0.0415}&0.0554\small{±0.0084}\\
RoBERTa MC Dropout     &0.8056\small{±0.0056}&0.8724\small{±0.0023}&0.3340\small{±0.0452}&0.1074\small{±0.0165}& 0.6217\small{±0.0034}&0.4907\small{±0.0082}&0.3340\small{±0.0452}&0.0604\small{±0.0060}\\
RoBERTa w/ BNN  &0.7992\small{±0.0022}&0.8699\small{±0.0014}&0.3069\small{±0.0010}&0.1228\small{±0.0021}&0.6227\small{±0.0002}&0.4686\small{±0.0056} &0.3069\small{±0.0010}&0.0881\small{±0.0002}\\
RoBERTa w/ \texttt{CUE} (Ours)          &0.8075\small{±0.0087}&0.8744\small{±0.0045}&\underline{0.6077}\small{±0.0595}&\textbf{0.0465}\small{±0.0075}&0.6255\small{±0.0005}&0.4540\small{±0.0013}&\underline{0.6077}\small{±0.0595}&\textbf{0.0316}\small{±0.0024}\\
\midrule
                 & \multicolumn{4}{c}{Emotion}                                                                 & \multicolumn{4}{c}{MultiNLI}                                                                     \\
\cmidrule(lr){2-5} \cmidrule(lr){6-9} 
Model             &  Acc    &  F1  & $\mathcal{H}$  & ECE$\downarrow$    &  Acc    &  F1  & $\mathcal{H}$  & ECE$\downarrow$   \\ \midrule
\midrule
ALBERT (11M)          &0.9284\small{±0.0037}&0.8862\small{±0.0031}&0.8862\small{±0.0031}&0.0348\small{±0.0059} &0.8362\small{±0.0018}&0.8358\small{±0.0018}  &0.8358\small{±0.0018}&0.0465\small{±0.0036} \\  
ALBERT Label Smoothing&0.9310\small{±0.0023}&0.8897\small{±0.0043} &\underline{0.8897}\small{±0.0043}&\textbf{0.0231}\small{±0.0018} &0.8327\small{±0.0020}&0.8317\small{±0.0020}   &0.8317\small{±0.0020}&0.0364\small{±0.0034} \\  
ALBERT MC Dropout     &0.9331\small{±0.0033}&0.8927\small{±0.0036}&0.8827\small{±0.0036}&0.0326\small{±0.0032} &0.8367\small{±0.0013}&0.8361\small{±0.0019}  &\underline{0.8361}\small{±0.0019}&0.0470\small{±0.0039} \\   
ALBERT w/ BNN         &0.9265\small{±0.0008}&0.8862\small{±0.0015}&0.8862\small{±0.0015}&0.0411\small{±0.0007} &0.8339\small{±0.0001}&0.8338\small{±0.0001}  &0.8338\small{±0.0001}&0.0527\small{±0.0001} \\  
ALBERT w/ \texttt{CUE}  (Ours)       &0.9269\small{±0.0020}&0.8897\small{±0.0044} &\underline{0.8897}\small{±0.0044}&0.0282\small{±0.0018}&0.8331\small{±0.0003}&0.8329\small{±0.0003}&0.8329\small{±0.0003}&\textbf{0.0338}\small{±0.0007}	 \\   \midrule
DistilBERT (66M)         &0.9287\small{±0.0031}&0.8886\small{±0.0062} &0.0441\small{±0.0044}&0.0393\small{±0.0030} &0.8067\small{±0.0014}	&0.8059\small{±0.0012} &0.3737\small{±0.0058}&0.0376\small{±0.0035} \\   
DistilBERT Label Smoothing&0.9264\small{±0.0031}&0.8841\small{±0.0056}&0.0716\small{±0.0051}&0.0353\small{±0.0020} &0.8049\small{±0.0012}	&0.8040\small{±0.0011} &0.3994\small{±0.0070}&0.0319\small{±0.0044} \\  
DistilBERT MC Dropout     &0.9298\small{±0.0018}&0.8886\small{±0.0025}&0.0432\small{±0.0051}&0.0388\small{±0.0020} &0.8066\small{±0.0020}	&0.8058\small{±0.0019} &0.3734\small{±0.0121}&0.0383\small{±0.0066} \\  
DistilBERT w/ BNN         &0.9315\small{±0.0008}&0.8931\small{±0.0009}&0.0413\small{±0.0001}&0.0406\small{±0.0000} &0.8059\small{±0.0002}	&0.8052\small{±0.0002} &0.3673\small{±0.0000}&0.0424\small{±0.0002}\\ 
DistilBERT w/ \texttt{CUE}  (Ours)       &0.9295\small{±0.0010}&0.8911\small{±0.0011} &\underline{0.0900}\small{±0.0002}&\textbf{0.0265}\small{±0.0005}&0.8058\small{±0.0003}	&0.8051\small{±0.0003}&\underline{0.4600}\small{±0.0022}&\textbf{0.0229}\small{±0.0005} \\ \midrule
BERT (110M)         & 0.9296\small{±0.0030}	&0.8871\small{±0.0057}	&0.0523\small{±0.0011}&0.0335\small{±0.0020}&0.8286\small{±0.0029}&0.8281\small{±0.0027} &0.3361\small{±0.0071}&0.0321\small{±0.0033} \\ 
BERT Label Smoothing& 0.9295\small{±0.0042}	&0.8862\small{±0.0074}	&0.0759\small{±0.0024}&\textbf{0.0289}\small{±0.0061}&0.8265\small{±0.0016}&0.8261\small{±0.0017}&0.3513\small{±0.0116}&0.0317\small{±0.0043} \\ 
BERT MC Dropout     & 0.9285\small{±0.0027}	&0.8872\small{±0.0048}	&0.0527\small{±0.0026}&0.0325\small{±0.0037}&0.8287\small{±0.0030}&0.8281\small{±0.0028} &0.3363\small{±0.0071}&0.0315\small{±0.0032} \\ 
BERT w/ BNN         & 0.9274\small{±0.0008}	&0.8853\small{±0.0011}	&0.0497\small{±0.0001}&0.0402\small{±0.0006}&0.3469\small{±0.0194}&0.1862\small{±0.0179}&\underline{0.9245}\small{±0.0835}&0.1456\small{±0.1111} \\ 
BERT w/ \texttt{CUE}  (Ours)       & 0.9259\small{±0.0009}&	0.8850\small{±0.0015}&\underline{0.1031}\small{±0.0082}&\textbf{0.0289}\small{±0.0043}&0.8283\small{±0.0005}&0.8277\small{±0.0005}&0.3665\small{±0.0030}&\textbf{0.0262}\small{±0.0021}\\ \midrule
RoBERTa (125M)         & 0.9279\small{±0.0033}&0.8821\small{±0.0062}&0.0448\small{±0.0043}&0.0384\small{±0.0050}&0.8569\small{±0.0043}&0.8563\small{±0.0044}&0.2628\small{±0.0127}&0.0368\small{±0.0074} \\
RoBERTa Label Smoothing& 0.9301\small{±0.0024}&0.8896\small{±0.0051}&0.0675\small{±0.0026}&0.0341\small{±0.0049}&0.8551\small{±0.0022}&0.8546\small{±0.0020}&0.3029\small{±0.0082}&\textbf{0.0255}\small{±0.0033} \\
RoBERTa MC Dropout     & 0.9305\small{±0.0034}&0.8919\small{±0.0059}&0.0514\small{±0.0068}&0.0315\small{±0.0046}&0.8586\small{±0.0062}&0.8581\small{±0.0061}&0.2516\small{±0.0309}&0.0403\small{±0.0112}\\
RoBERTa w/ BNN         & 0.9290\small{±0.0006}&0.8923\small{±0.0005}&0.0478\small{±0.0002}&0.0412\small{±0.0007}&0.8528\small{±0.0001}&0.8527\small{±0.0001}&0.2620\small{±0.0000}&0.0416\small{±0.0002}\\
RoBERTa w/ \texttt{CUE} (Ours)        & 0.9286\small{±0.0005}&0.8927\small{±0.0008}&\underline{0.0944}\small{±0.0004}&\textbf{0.0313}\small{±0.0033}&0.8526\small{±0.0003}&0.8526\small{±0.0003}&\underline{0.3179}\small{±0.0003}&0.0262\small{±0.0004}\\
\bottomrule
\end{tabular}}

\caption{Results for \texttt{CUE} compare against baseline methods on four language models trained on four datasets. Values shown in parentheses indicate the model size. The highest entropy are underlined and the lowest ECE values are in bold. }
\label{table:results}
\end{table*}

In this subsection, we discuss how to quantify the prediction uncertainty that caused by the input features based on the latent space perturbation in \textsection{\ref{sec:counterfactual}}. The discussion is built on an inner product space defined by our noise generation methods. 
During the inference stage, we identify the possible feature that caused predictive uncertainty by our proposed Uncertain Feature Identification (UFI) algorithm\footnote{We provide the UFI algorithm implementation in Supplementary Material Section 2.}. For a given input, we can retrieve three different representations from the \texttt{CUE} framework, the original PLM-encoded representation $\bm{e}_i$, the reconstructed representation $\bm{e}'_i$, and the difference between two representation, $\Delta e_i = \bm{e}'_i - \bm{e}_i$.
The reconstructed representation $\bm{e}'_i$ can be rewritten as the weighted sum of each latent dimension from $\bm{z}_i$, where the weight is given by the decoder:
\begin{equation}
\small
    \bm{e}'_i = \Sigma_{d=1}^{\rm dim} {q_{\phi}({\bm{z}_i}_d|\bm{e}_i)} \cdot {\bm{r}_{z_i}}_d, \quad {\bm{r}_{z_i}}_d = \mu_{\theta}({\bm{z}_i}_d),
    \label{eq:GMPR}
\end{equation}
\noindent where $\rm dim$ is the size of the latent space and ${\bm{r}_{z_i}}_d$ denotes the representation generated via the $d$-th dimension's code corresponding to the latent vector $\bm{z}_i$ from the decoder. As mentioned in \textsection{\ref{sec:counterfactual}}, $\Delta \mathcal{H}$ 
is proportional to the reconstruction error $||\bm{e}'_i-\bm{e}_i||^2$. We thus use the norm (calculated as the inner product) of the reconstruction error, $||\Delta \bm{e}_i||^2 = \langle \Delta \bm{e}_i, \Delta \bm{e}_i \rangle $, to measure the entropy change as: 
{\small
\begin{align}
\label{eq:detalNorm}
    \langle \Delta \bm{e}_i, \Delta \bm{e}_i \rangle 
    =&  \langle \Sigma_{d=1}^{\rm dim} {q_{\phi}({\bm{z}_i}_d|\bm{e}_i)} \cdot {\bm{r}_{z_i}}_d - \bm{e}_i, \Delta \bm{e}_i \rangle \notag \\
    =& \langle \Sigma_{d=1}^{\rm dim} {q_{\phi}({\bm{z}_i}_d|\bm{e}_i)} \cdot {\bm{r}_{z_i}}_d, \Delta \bm{e}_i \rangle  - \langle \bm{e}_i, \Delta \bm{e}_i \rangle, 
\end{align}}
\noindent where $\langle \cdot,\cdot \rangle$ denotes the inner product. 
In the first line of Eq. (\ref{eq:detalNorm}), we substitute the first $\Delta \bm{e}_i$ with $\bm{e}'_i -\bm{e}_i$, and further substitute $\bm{e}'_i$ with Eq. (\ref{eq:GMPR}). When determining the relative importance of each latent dimension with respect to the predictive entropy change, $\langle \bm{e}_i, \Delta \bm{e}_i\rangle$ can be ignored as it is the same for all latent dimensions. Therefore, the inner product of $\langle {\bm{r}_{z_i}}_d, \Delta \bm{e}_i \rangle$, which dominates the norm value of $||\Delta \bm{e}_i||$ in the $d$-th dimension can be used to measure predictive uncertainty caused by each dimension from the latent space $\bm{r}_{z_i}$, and thus determine each dimension's importance.

On the other hand, the input text representation $\bm{e}_i$ output by the PLM at layer-$L$ can be written as a Softmax-based weighted sum of each token's representation from the previous layer $L-1$ by\footnote{Note that all representations in the RHS are from Layer $L-1$. We drop the superscript $L-1$ to simplify the notations.}:
{\small
\begin{align}
    \bm{e}_i^L  &= \Sigma_{j=1}^n {\rm Softmax}(\left \langle \bm{e}_i, {\bm{e}_{i}}_j \rangle\right ) \cdot {\bm{e}_{i}}_j \nonumber\\
    &\propto \Sigma_{j=1}^n {\rm exp}(\langle \bm{e}_i,{\bm{e}_{i}}_j \rangle) \cdot {\bm{e}_{i}}_j,
\end{align}}
\noindent where ${\bm{e}_{i}}_j$ denotes the representation of the $j$-th input token. We assume that the adjacent layers in the transformer share similar representations. Then, $\bm{e}_i$ at layer $L$ is: 
{\small
\begin{align}
    \bm{e}_i^L \propto& \Sigma_{j=1}^n {\rm exp}(\langle \Sigma_{d=1}^{\rm dim} {q_{\phi}({\bm{z}_i}_d|\bm{e}_i)} \cdot {\bm{r}_{z_i}}_d,{\bm{e}_{i}}_j \rangle) \cdot {\bm{e}_{i}}_j \notag \\
    \propto & \Sigma_{j=1}^n ( \Sigma_{d=1}^{\rm dim} {q_{\phi}({\bm{z}_i}_d|\bm{e}_i)} \cdot \langle {\bm{r}_{z_i}}_d,{\bm{e}_{i}}_j \rangle) \cdot {\bm{e}_{i}}_j 
\end{align}}
Therefore, the influence on prediction uncertainty changes $\Delta \mathcal{H}$ of the $j$-th token is decided by the generative probability of the encoder and the inner product $\langle {\bm{r}_{z_i}}_d,{\bm{e}_{i}}_j \rangle$. However, seeking the optimal ${\bm{z}_i}_d$, by minimizing the reconstruction loss Eq. (\ref{eq:reconstruction}), is a typical Knapsack problem, which is an NP-complete problem. Hence, intuitively, we use greedy search to find a locally optimal solution by identifying the most influential latent dimensions of $\bm{z}_i$ first and then estimating the influential score for each token. 

\begin{figure*}[ht]
\centering
\subfloat[BERT \texttt{CUE} on CoLA.]{
    \includegraphics[width=\columnwidth]{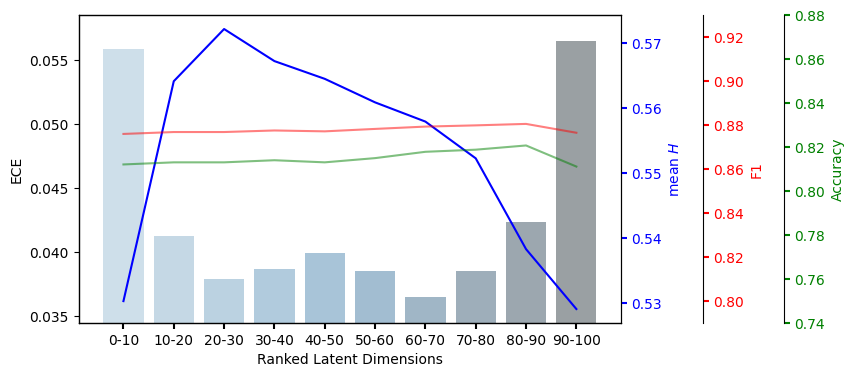}}
\subfloat[BERT \texttt{CUE} on GoEmotions.]{
    \includegraphics[width=\columnwidth]{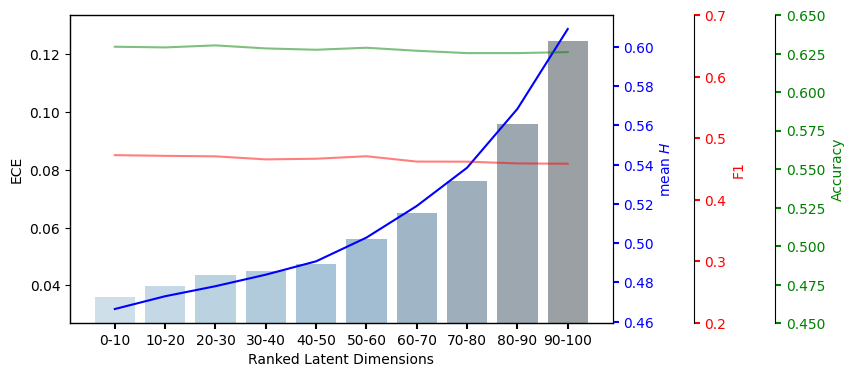}}\\
\subfloat[BERT \texttt{CUE} on Emotion.]{
    \includegraphics[width=\columnwidth]{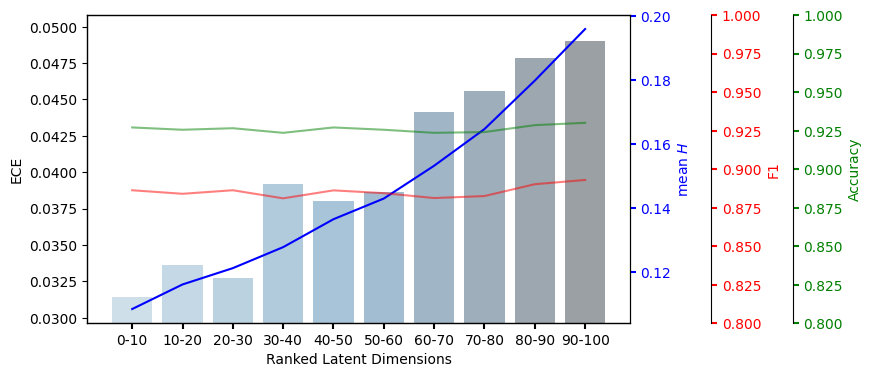}}
\subfloat[BERT \texttt{CUE} on MultiNLI.]{
    \includegraphics[width=\columnwidth]{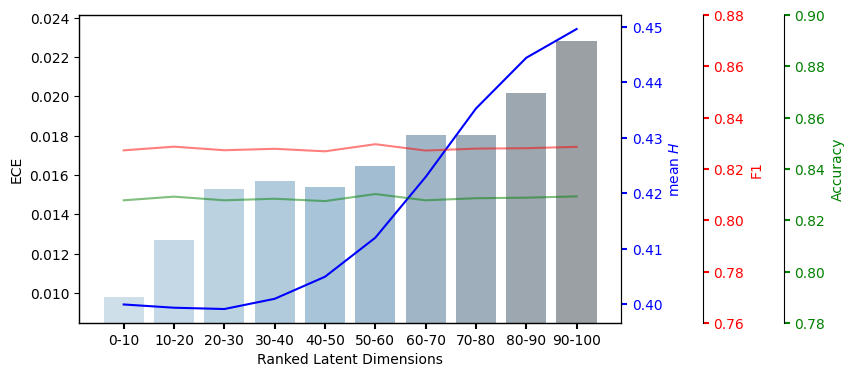}}
\caption{Evaluation results by removing latent dimensions. The $x$-axis represents the index of \textbf{removed} dimensions ranked by their relevance to $\Delta \bm{e}_i$, smaller index number indicates higher relevance. Histograms show the ECE scores after removing the corresponding latent dimensions. The blue curve shows the predictive entropy. The green and red curves show classification accuracy and F1, respectively.}
    \label{fig:latent_ablation}
\end{figure*}

\section{Experiments} \label{sec:experiments}
We first present the experimental setup followed by evaluation results.

\paragraph{Datasets}
We evaluate our proposed framework on four datasets for \emph{linguistic acceptability classification}, \emph{natural language inference}, and \emph{emotion classification}. 

\paragraph{Baselines}
We compare our method with three baselines: Label Smoothing \citep{Gupta_Kvernadze_Srikumar_2021}, MC Dropout \citep{monte_carlo_dropout} and Bayesian Neural Network (BNN). Label Smoothing and MC Dropout are implemented in PLMs and directly fine-tuned on the target datasets. The BNN works as a plug-in component, same as \texttt{CUE}, for which the base PLM encoding and the classification layer are firstly fine-tuned and then parameters are frozen for the plug-in layer training. 

\textbf{Evaluation Metrics}
Accuracy (Acc), macro-averaged F1 (F1), average entropy ($\mathcal{H}$), and Expected Calibration Error (ECE) are used as metrics for classification performance, uncertainty and model calibration measurement. 

More details on dataset statistics, baseline setup, evaluation metrics and hyperparameter settings are in Supplementary Material Section 3.


\newcommand\blue[1]{\textcolor{blue}{\emph{#1}}}
\newcommand\red[1]{\textcolor{red}{#1}}

\begin{table*}[htb]
\begin{center}
\resizebox{\linewidth}{!}{
\begin{tabular}{p{0.95\textwidth}p{0.15\textwidth}p{0.1\textwidth}}
\toprule
\textbf{Examples} & \textbf{Predicted} & \textbf{True} \\
\cmidrule(lr){1-1} \cmidrule(lr){2-2} \cmidrule(lr){3-3}
\multicolumn{3}{c}{\textbf{GoEmotions}} \\ \midrule
Despite having lived here for 10 years, I've never been to portillos, and given this, it's \blue{somewhat unlikely} I start going now...   & Disapproval 0.42$\rightarrow$ 0.35 & Neutral  \\
Somehow I got banned for replying to a troll. The mods over there have \blue{itchy trigger fingers}. & Disappointment\newline 0.28 $\rightarrow$ 0.19 & Disapproval \\
Boundaries. Have some boundaries. \blue{Say no}. \blue{Don't go}. This is frustrating to read, honestly. Don't do things that you hate doing. & Disgust \newline 0.34 $\rightarrow$ 0.25 & Fear \\\midrule
\multicolumn{3}{c}{\textbf{Emotion}} \\
\midrule
I were to \blue{go overseas or cross the border} then I become a foreigner and will feel that way but never in my \blue{beloved land}. & Joy\newline 0.51 $\rightarrow$ 0.43 & Love \\
I started feeling a little stressed about leaving on time and making sure we \blue{got the getting ready pictures} I wanted but everything seemed to work out perfectly. & Sadness\newline 0.59 $\rightarrow$ 0.40 & Anger \\
I \blue{wont} lie this week has been \blue{abit of a difficult} week for me ive been feeling very stressed and anxious this week plus i think im coming down with the flu but it \blue{has definately} helped me to appreciate the little things. & Sadness\newline  0.64 $\rightarrow$ 0.44 & Anger \\\midrule
\multicolumn{3}{c}{\textbf{MultiNLI}} \\
\midrule
\textbf{P:} There are no shares of a stock that might someday come back, just piles of options \blue{as worthless as} those shares of cook's american business alliance. & Neutral\newline 0.43 $\rightarrow$ 0.40 & Contradiction  \\
\textbf{H:} Cook's american business alliance caused shares of stock to come back. & & \\
\textbf{P:} Until all \blue{members} of our \blue{society} are afforded that access, this promise of our government will continue to be \blue{unfulfilled}. & Entailment\newline 0.48 $\rightarrow$ 0.43 & Neutral\\
\textbf{H:} the government is flawed and \blue{unfulfilled}. & &\\
\textbf{P:} Here you`ll find the finest leather goods and of - the - moment fashions from all the predictable high - \blue{priests} (valentino, armani, versace, gucci, missoni, etc.). A number of \blue{classic} men's clothing meccas such as cucci (with a c), brioni, and battistoni are still \blue{going strong}. & Contradiction\newline 0.35 $\rightarrow$ 0.33 & Entailment \\
\textbf{H:} You will find only the highest quality goods, be they \blue{high - fashion icons} or top - \blue{notch designer clothing} here. & &\\
\bottomrule
\end{tabular}}
\caption{Visualisation of token-level uncertainty interpretation. The `\textbf{Predicted}' column denotes the incorrect predictions made by the original fine-tuned model. Values below each predicted label denotes the predictive probability change after applying our framework. The `\textbf{True}' column denotes the gold-standard class labels. Italic text highlighted in blue are word tokens identified by UFI Algorithm that cause predictive uncertainty.}
\label{tab:sen_examples}
\end{center}
\end{table*}

\subsection{Overall Comparison}

Table \ref{table:results} presents the performance of methods with four state-of-the-art PLMs, namely,  BERT \citep{bert}, ALBERT \citep{albert}, DistilBERT \citep{distilbert} and RoBERTa \citep{roberta}, as backbones. 
Our framework with a plug-in \texttt{CUE} module obtains the lowest ECE scores and highest average predictive entropy on all tasks and with different base model choices while maintaining a comparable level of Acc/F1 scores as the original model. Although the BNN model achieves the highest entropy with BERT on MultiNLI and CoLA dataset, we can observe a significant drop in its Acc/F1 scores. This indicates the BNN encoder hardly generates reliable perturbations that maintain predicted labels unchanged.
Interestingly, while the classification performance of all compared models shows slight degradation with the injection of uncertainty into the PLMs, our framework achieves steady accuracy gains on the CoLA dataset. 
\subsection{Effectiveness of the Uncertainty Feature Identification Algorithm} 
\label{sec:effectiveness_experiments}
\paragraph{Results with Latent Dimension Removal}\label{sec:label_removal}
As presented in \textsection{\ref{sec:inputFeatureIdentification}}, we can use the \texttt{CUE}'s reconstruction difference $\Delta{\bm{e}_i}$, to disentangle the most influential latent dimensions ${\bm{z}_i}_d$s which cause predictive uncertainty. Since each latent dimension is associated with an influential score, we can sort the latent dimensions accordingly. We speculate that by removing latent dimensions with higher influence scores, we should be able to observe a reduction in predictive uncertainty. As shown in Figure \ref{fig:latent_ablation}, we visualise the evaluation results by removing latent dimensions from $\bm{z}_i$ according to their relevance to $\Delta \bm{e}_i$ (the rank is shown on the $x$-axis) on BERT models. 
In our experiments, the latent vector $\bm{z}_i$ has 100 dimensions, we thus sort them into 10 bins in descending order based on their influential scores. In practice, the latent dimension removal is achieved by assigning $0$ as the value of the dimension on $\bm{z}_i$ to create a modified latent variable $\bm{z}_i'$, new prediction is made with $\hat{y}_i' = g_{\bm{\eta}}(p_{\bm{\theta}}(\bm{z}_i'))$.  

We can observe a remarkable increasing trend of ECE (the histograms) and average entropy (the blue curve) when removing the most influential latent dimensions of $\bm{z}_i$ on GoEmotions, Emotion and MultiNLI datasets. This indicates the top-ranked dimensions (smaller index number) contribute more to increase the predictive uncertainty and reducing the overconfidence prediction, 
while lower-ranked dimensions have less effect. Therefore, we can select the appropriate threshold for each dataset during token-level uncertainty identification by observing the trend of ECE and entropy.
We also notice that across all datasets, removing any latent dimensions does not affect much the classification accuracy (the green curve) and macro-F1 (the red curve). However, we did not observe a similar trend of ECE and entropy on the CoLA dataset. We suspect this is due to a relatively simple setup in CoLA as it is only a binary classification task. For datasets with more classes, such as GoEmotion with 27 classes, the trend of ECE with latent dimension removal becomes more obvious. 
We also performed the same analysis and observed similar phenomena on other PLMs, DistilBERT, ALBERT and RoBERTa, in Supplementary Material Section 4.1.  

\paragraph{Case Study of Token-Level Uncertainty Identification}

In this subsection, we demonstrate the effectiveness of our uncertainty identification algorithm by visualising the tokens that our framework finds contributing to predictive uncertainty. We present several examples in which our framework reduces overconfident predictions 
in Table \ref{tab:sen_examples}. We only show the results with BERT as the base model due to page limits. 
Tokens coloured in blue are the influential tokens\footnote{For words split into subword tokens, we take the average importance score of the constituent subword tokens.} identified by the UFI Algorithm.

For emotion classification, we found classifiers tend to be confused by idioms or phrases carrying emotions different from the true emotion labels. For example, the second sentence in GoEmotion contains a metaphorical phrase, `\emph{itchy trigger finger}', making it a tricky case for emotion classification. We conducted additional experiments in which we substituted the phrase "itchy trigger finger" with either the \texttt{\small{[MASK]}} token or commonly used words to express the same meaning. In both cases, the model uncertainty is reduced by replacing the original phrase with the mask tokens leading to label switching. Replacing the identified phrase with more commonly-used words increases the predictive probability and leads to a more confident prediction. These results verify the validity of our approach for identifying words/phrases causing predictive uncertainties. The first and last sentences in GoEmotion and also the last sentence in Emotion contain phrases which are somewhat more closely related to the incorrectly predicted labels than the true labels, confusing the classifier to generate wrong predictions. 
Tokenisation may also cause a problem. For example, the word `\emph{beloved}' in the first sentence in Emotion is split into three parts after tokenisation, making it difficult for the classifier to recognise 
the `Love' emotion. For the natural language inference task, we found classifiers tend to make overconfident predictions when the same words are found in both premise and hypothesis. For examples, the second instance in MultiNLI has the word `\emph{unfulfilled}' in both its premise and hypothesis. This leads to the wrong prediction of `Entailment'. In the last instance, 
the classifier misunderstood that the `\emph{classic men's clothing}' contradicts with `\emph{high - fashion icons}' and thus failed to recognise the `Entailment' relation. 
Nevertheless, in all these cases, our proposed framework managed to increase the predictive entropy by reducing the confidence of predictions, alleviating the overconfidence problem.

We provide further experimental results and the ablation study, including stability of various additional training loss terms and latent space orthogonality in Supplementary Material Section 4.2.
\section{Conclusion}
In this paper, we have proposed a new framework \texttt{CUE} for uncertainty interpretation of PLM classifiers. 
By comparing our method with previous solutions, we show that \texttt{CUE} can achieve lower expected calibration errors across four datasets. In some cases, it can also mitigate the confidence of previously wrong predictions. Further experiments and case studies demonstrate \texttt{CUE} is effective in identifying tokens/latent dimensions that could potentially cause predictive uncertainty. 
Our work sheds light on a new direction of uncertainty interpretation for PLMs in various NLP tasks.

\begin{acknowledgements} 
This work was supported in part by the UK Engineering and Physical Sciences Research Council (grant no. EP/T017112/2, EP/V048597/1, EP/X019063/1). YH is supported by a Turing AI Fellowship funded by the UK Research and Innovation (grant no. EP/V020579/2). The authors would like to thank Yuxiang Zhou, Hanqi Yan and Xingwei Tan for their invaluable feedback on this paper. 
\end{acknowledgements}

\bibliography{li_516}

\end{document}


\onecolumn 
\maketitle

\section{Derivations}

\subsection{Decomposition of the Predictive Uncertainty}
We show how we decompose the Mean Squared Error (MSE) based predictive uncertainty into the epistemic uncertainty and the aleatoric uncertainty mentioned in \textsection{3.2}. 
\begin{align}
    \mathbb{E}\big[(y_i - \hat{y}_i)^2\big] 
    &= \mathbb{E}\big[(y_i - \mathbb{E}[y] + \mathbb{E}[y] - \hat{y}_i)^2\big] \notag\\
    &=\mathbb{E}\big[(y_i - \mathbb{E}[y])^2] + \mathbb{E}[(\mathbb{E}[y] - \hat{y}_i)^2] + 2\mathbb{E}[(y_i - \mathbb{E}[y])(\mathbb{E}[y] - \hat{y}_i)] \notag\\
    &= \mathbb{E}[(y_i - \mathbb{E}[y])^2] + \mathbb{E}[(\mathbb{E}[y] - \hat{y}_i)^2] + 2(\mathbb{E}[{y}] - \mathbb{E}[y])(\mathbb{E}[y] - \mathbb{E}[\hat{y}_i]) \notag\\ 
    & = \underbrace{\mathbb{E}[(y_i - \mathbb{E}[y])^2]}_{\rm aleatoric\,uncertainty} + \underbrace{\mathbb{E}[(\mathbb{E}[y] - \hat{y}_i)^2]}_{\rm epistemic\,uncertainty} \notag
\end{align}
Here, $\mathbb{E}[y]$ denotes the expectation of the ground truth label distribution. Since the first term, $\mathbb{E}[(y_i - \mathbb{E}[y])^2]$, contains the observed $y_i$, it can be defined as the aleatoric uncertainty. The second term represents the epistemic uncertainty since it contains the predicted $\hat{y}_i$.

\subsection{Interpreting Entropy Change with Reconstruction Difference}
In this subsection, we provide detailed derivation corresponding to the predictive entropy upper-bound mentioned in \textsection{4.1}.

In our learning objective, we aim to estimate the uncertainty by adding the noise to increase the predictive entropy while keeping the classification results unchanged. For a Softmax-based classifier, the predictive uncertainty reaches the maximum when probabilities are uniformly distributed (i.e. when the predictive class probability of any of the $K$ classes is $\frac{1}{K}$).  
Then, we have 
\begin{equation}
    0 \leq \mathcal{H}_{\bm{e}_i}(\hat{y}_i) \leq \mathcal{H}_{\bm{e}'_i}(\hat{y}'_i) \leq {\rm log}K  \notag
\end{equation}
Assuming that the difference between text representation $\bm{e}_i$ and the reconstructed representation $\bm{e}'_i$ is $\bm{u}$, $\bm{u} = \bm{e}'_i - \bm{e}_i$, and the prediction is obtained from the Softmax function. Let $\bm{U}$ be the maximum distance of $\bm{e}'_i-\bm{e}_i$ that causes the new $\bm{e}'_i$ confuse the classifier (i.e. when the predictive class probability equals to $\frac{1}{K}$). Then, according to the Jensen inequality, the prediction is bounded by: 
\begin{align}
\notag
    \hat{y}_i' &\leq t \cdot {\rm softmax}(\bm{e}_i) + (1-t){\rm softmax}(\bm{e}_i+\bm{U}) \\ \notag
    & =   t \cdot {\rm softmax}(\bm{e}_i) + (1-t)\frac{1}{K} \notag
\end{align}
So we can get $\bm{e}_i \leq \bm{e}_i' \leq (\bm{e}_i + \bm{U})$, let $0 \leq t \leq 1$ and $\mathcal{H}_{(\bm{e}_i+\bm{U})}(\hat{y}_i') = {\rm log}K$. Considering the convexity of entropy, we have:
\begin{small}
\begin{align}
    \Delta \mathcal{H} &= \mathcal{H}_{\bm{e}_i'}(\hat{y}_i') - \mathcal{H}_{\bm{e}_i}(\hat{y}_i)\notag\\ \notag
    &\leq - (t \cdot p(\hat{y}_i) +  \frac{(1-t)}{K}) {\rm log} (t \cdot p(\hat{y}_i) + \frac{(1-t)}{K}) + p(\hat{y}_i){\rm log}p(\hat{y}_i)\\ \notag
    & \leq - t \cdot p(\hat{y}_i) {\rm log} p(\hat{y}_i) - (1-t) \cdot \frac{1}{K} {\rm log} (\frac{1}{K}) + p(\hat{y}_i){\rm log}p(\hat{y}_i)\\ \notag
    & = (1-t)({\log} K - \mathcal{H}_{\bm{e}_i}(\hat{y}_i) ) \\\notag
    & = \frac{||\bm{e}_i'-\bm{e}_i||^2 \cdot ({\log} K - \mathcal{H}_{\bm{e}_i}(\hat{y}_i) )}{||U||^2} \\
    &\propto ||\bm{e}_i'-\bm{e}_i||^2   \notag
\label{eq:entropy}
\end{align}
\end{small}

The above derivation demonstrates the generated perturbation can guarantee an upper bound of predictive entropy difference $\Delta\mathcal{H}$, and the variation of the entropy $\Delta\mathcal{H}$ is proportional to the reconstruction error $||\bm{e}'_i-\bm{e}_i||^2$, which thus can be used to interpret the predictive uncertainty.

\section{Uncertain Feature Identification Algorithm}
In this section, we provide the detailed implementation of the Uncertain Feature Identification algorithm built in the \texttt{CUE} framework corresponding to \textsection{4.2}. Intuitively, we use greedy search to find a locally optimal solution by identifying the most influential latent dimensions of $\bm{z}_i$ first and then estimating the influential score for each token (see in Algorithm \ref{alg:UFI_alg}). That is, we can identify input tokens that are most similar to the influential representation vector ${\bm{r}_{z_i}}_d$ as the ones which cause predictive uncertainty by the inner product in the metric space. More concretely, assuming the PLM-encoded representation for token $j$ is ${\bm{e}_{i}}_j$, we can compute each token's importance score by ${\rm{token}^j}_{\rm score}= \langle {\bm{r}_{z_i}}_d,{\bm{e}_{i}}_j  \rangle$. By sorting ${\rm{token}^j}_{\rm score}$ in descending order, we can identify input tokens that cause predictive uncertainty.

The identification of the source of the uncertainty highly relies on the influence of latent dimensions. In practice, we use a threshold $\alpha$ to select the most similar dimensions of $\Delta \bm{e}_i$ to construct a combination of the most influential uncertain representation ${\bm{r}_{z_i}}_D$. The threshold $\alpha$ can be defined with the help of the average entropy curve and ECE histograms from the dimension importance analysis described in \textsection{5.2}.

\begin{algorithm}[h]
\small
\caption{Uncertain Feature Identification}
\label{alg:UFI_alg}
\textbf{Input:} original text representation $\bm{e}_i$, reconstruct text representation $\bm{e}'_i$, \texttt{CUE} decoder $\mu_{\bm{\theta}}$, token representations $\{{\bm{e}_{i}}_1,{\bm{e}_{i}}_2,\cdots,{\bm{e}_{i}}_n\}$, tokens $\{\rm{token}_1,\rm{token}_2,\cdots,\rm{token}_n\}$, threshold $\alpha$.
\normalsize
\begin{algorithmic}
\STATE $\Delta \bm{e}_i = \bm{e}'_i - \bm{e}_i$
\FOR {the $d$th dimension ${\bm{z}_{i}}_d$ in $\bm{z}_i$}
    \STATE ${\bm{r}_{z_i}}_d = \mu_{\theta}({\bm{z}_{i}}_d)$ 
    \STATE ${\rm dim}^{d}_{\rm score}= \langle \Delta \bm{e}_i, {\bm{r}_{z_i}}_d \rangle$
\ENDFOR
\FOR {$\mbox{sort}({\bm{r}_{z_i}}_d, key=\phi({\rm dim}^{d}_{\rm score}, {\bm{r}_{z_i}}_d))[:\alpha]$}
    \STATE ${\bm{r}_{z_i}}_D += {\bm{r}_{z_i}}_d$
\ENDFOR
\FOR {${\bm{e}_{i}}_j$ in $\{{\bm{e}_{i}}_1,{\bm{e}_{i}}_2,\cdots,{\bm{e}_{i}}_n\}$}
    \STATE ${\rm{token}^j}_{\rm score}=  \langle {\bm{r}_{z_i}}_D, {\bm{e}_{i}}_j  \rangle$
\ENDFOR
\STATE \textbf{return} $\mbox{sort}(\rm{token}_j, key=\phi({\rm{token}^j}_{\rm score}, \rm{token}_j))$
\end{algorithmic}
\end{algorithm}

\section{Experimental Setup}
In this section, we provide detailed dataset statistics, baseline setup, evaluation metrics and hyperparameter settings as mentioned in \textsection{5}.

\paragraph{Datasets}

We evaluate our proposed framework on four datasets for \emph{linguistic acceptability classification}, \emph{natural language inference}, and \emph{emotion classification}. The dataset statistics are shown in Table \ref{table:dataset}.

\begin{table}[h]
\centering
\begin{tabular}{lrrrr}
\toprule
Datasets &   CoLA & MultiNLI & Emotion & GoEmotions \\ \midrule
Classes  &  2 & 3 & 6 & 27  \\   \midrule
 Train   &   8,551   &  392,702     &   16,000 &   43,410 \\ 
 Dev     &   1,043   &    20,000    &   2,000  &   5,427  \\ 
 Test    &   1,043   &    20,000    &   2,000  &   5,426  \\ \midrule
Total    &   10,637  &    432,702   &   20,000 &   58,009 \\
\bottomrule
\end{tabular}
\caption{Statistic of the datasets.}
\label{table:dataset}
\end{table}
\noindent\underline{Linguistic Acceptability Classification}. The CoLA (Corpus of Linguistic Acceptability) 
\citep{cola} 
contains sentences 
annotated as \emph{grammatically acceptable} or \emph{not}.

\noindent\underline{Natural Language Inference}. The MultiNLI \citep{dataset_multinli} 
dataset contains annotations for relations of \emph{entailment}, \emph{contradiction}, and \emph{neutrality} between sentence pairs.

\noindent\underline{Emotion Classification}. 
The GoEmotions \citep{dataset_goemotions} dataset annotates Reddit comments with twenty-seven emotion labels (e.g., \emph{fear} and \emph{admiration}). 
The Emotion \citep{dataset_emotion} dataset classifies English tweets into six emotion classes (e.g., \emph{sadness} and \emph{joy}). Note that the GoEmotions dataset allows multi-label settings that a sentence can be annotated with more than one emotion label. In our experimental setup, we only focus on multi-class classification, and we thus filtered out those instances annotated with multiple labels in the GoEmotions.

\paragraph{Baselines}

We compare our method with the following baselines:

\noindent\underline{Label Smoothing} \citep{ Gupta_Kvernadze_Srikumar_2021} is commonly used to deal with overfitting 
when using cross-entropy loss on classification tasks. It aims to uniform the distribution of labels to encourage small logit gaps and has been shown effective in calibrating PLM-based classifiers. 

\noindent\underline{MC Dropout} \citep{monte_carlo_dropout} 
is an uncertainty estimation technique that performing multiple stochastic forward passes by randomly switching neurons off to generate ensemble of predictions. We follow the implementation of \citet{acl2022_uncertainty_transformers} in our experiments. 

\noindent\underline{Bayesian Neural Network (BNN)} \citep{8371683} assumes weights of neural networks are random variables with a prior distribution, is thus able to obtain more robust predictions by sampling the network weights during inference, and is often used for uncertainty estimation. Motivated by \citet{getting_clue}, we also implemented a BNN plug-in framework as a comparison with our \texttt{CUE} framework. Specifically, we use a Bayesian linear layer\footnote{\url{https://github.com/piEsposito/blitz-bayesian-deep-learning}} as the encoder and a linear layer as the decoder, and then insert them between the PLM-encoding layer and the classification layer, similar to the way we ensemble the \texttt{CUE}.


\paragraph{Evaluation Metrics}

We use accuracy (Acc) and macro-averaged F1 (F1) to evaluate the classification results, and Expected Calibration Error (ECE) \citep{desai-durrett-2020-calibration} calculated on predictive probabilities during inference to measure model calibration. 
For ECE implementation, we use the formula provided by \citet{softmax_uncertain} as follows:
\begin{small}
\begin{gather}
    \text{acc}(B_m) = \frac{1}{B_m} \sum_{i\in B_m} 1(\hat{y}_i = y_i), \quad\quad
    \text{conf}(B_m) = \frac{1}{B_m} \sum_{i\in B_m} \hat{p}_i, \notag\\
    \mbox{ECE} = \sum_{m=1}^M \frac{B_m}{n} |\text{acc}(B_m)-\text{conf}(B_m)| \notag
\end{gather}
\end{small}

Predictions of $n$ samples are grouped into $M$ interval bins and the accuracy is calculated for each bin. $B_m$ is the set of indices of samples that prediction confidence falls into the current interval bin. The ECE formula calculates the weighted average of the difference between the accuracy of each bin -  $\text{acc}(B_m)$ - and the average confidence - $\text{conf}(B_m)$ - within bin $B_m$. 
In our experiments, we set $B_m = 9$.

\paragraph{Hyperparameters Settings} 

We adopted the Pytorch-Transformers package\footnote{\url{https://github.com/huggingface/pytorch-transformers}} for the implementation of all our Transformer-based language models. For each model, we chose its corresponding base model with the following parameter size: ALBERT-base-v2 (11M), distilBERT-base-uncased (66M), BERT-base-uncased (110M), and RoBERTa-base (125M). 
We fine-tuned all these base models for 20 epochs with a batch size of 16 on each target dataset as compared to base models. For the Label Smoothing and the MC Dropout baseline, the frameworks directly modified the PLM-based models and were finetuned together with the PLM for 20 epochs with a batch size of 16. For the BNN and our \texttt{CUE} plug-in methods, we first fine-tuned the base models for 20 epochs and then froze the PLM encoding and classifier parameters and fine-tuned the BNN and \texttt{CUE} module for a further 50 epochs (with batch sizes as 16 for both modules).
A learning rate of $2e-5$ and the early stop strategy have been applied to all the training. 
Each model has been trained 5 times with different random seeds. 
For each model, we report the mean and standard deviation of the evaluation results obtained by the five trained models on test sets. 

\section{Further Experimental Results}\label{sec:futher_experimental_results}


\begin{figure}[h!]
\centering
\subfloat[ALBERT \texttt{CUE} on CoLA.]{
  \centering
  \includegraphics[width=0.35\columnwidth]{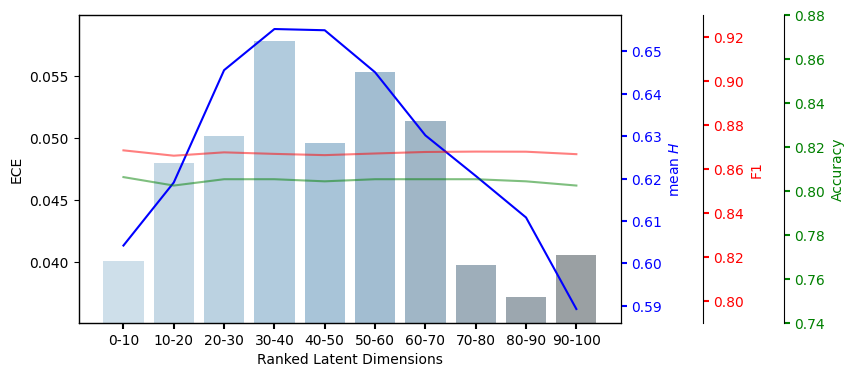}}
\subfloat[ALBERT \texttt{CUE} on GoEmotions.]{
  \centering
  \includegraphics[width=0.35\columnwidth]{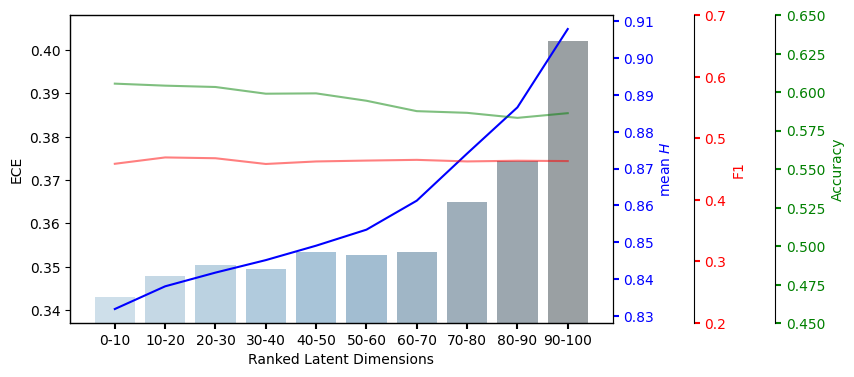}}\\
\subfloat[ALBERT \texttt{CUE} on Emotion.]{
  \centering
  \includegraphics[width=0.35\columnwidth]{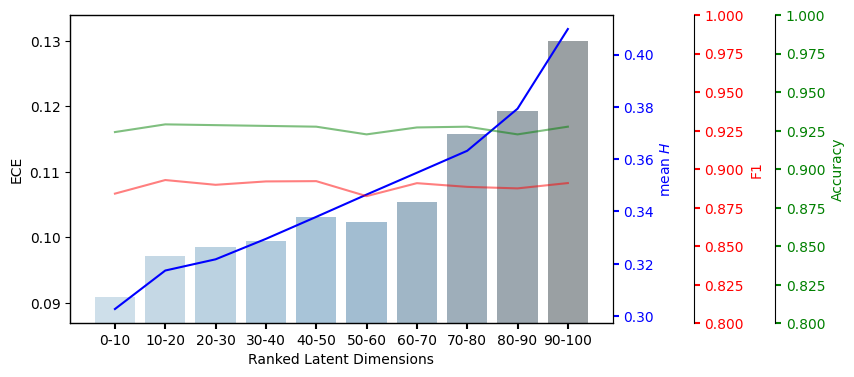}}
\subfloat[ALBERT \texttt{CUE} on MultiNLI.]{
  \includegraphics[width=0.35\columnwidth]{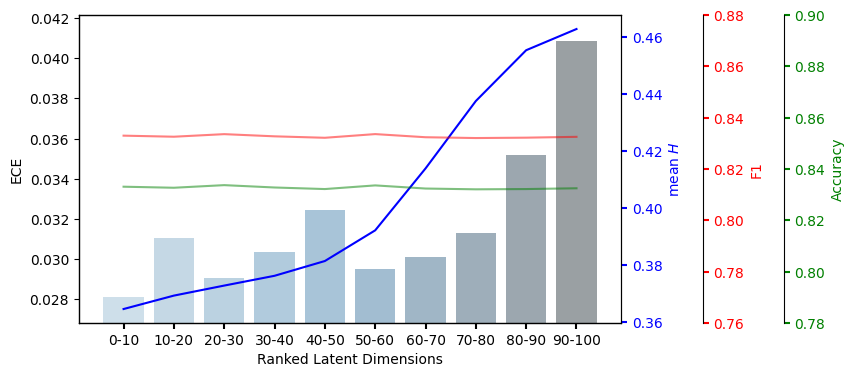}}\\
\subfloat[DistilBERT \texttt{CUE} on CoLA.]{
  \includegraphics[width=0.35\columnwidth]{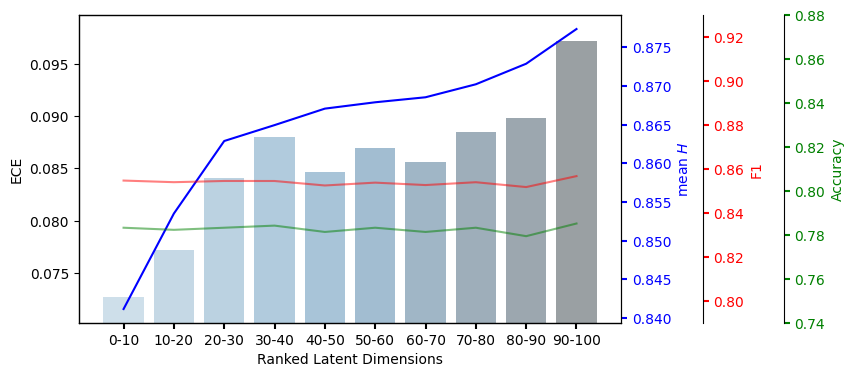}}
\subfloat[DistilBERT \texttt{CUE} on GoEmotions.]{
  \includegraphics[width=0.35\columnwidth]{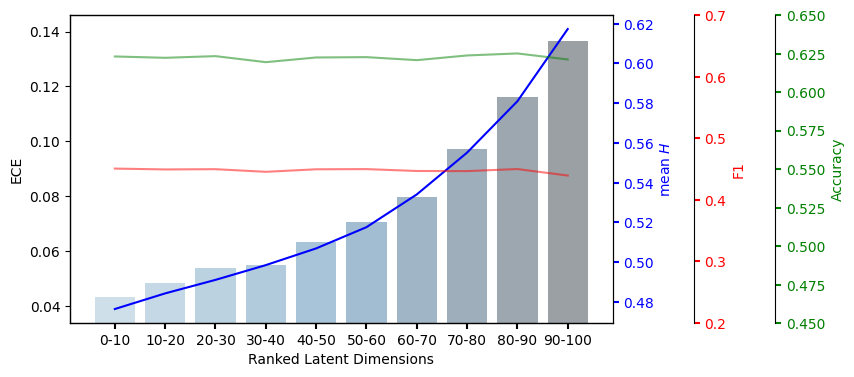}}\\
\subfloat[DistilBERT \texttt{CUE} on Emotion.]{
  \includegraphics[width=0.35\columnwidth]{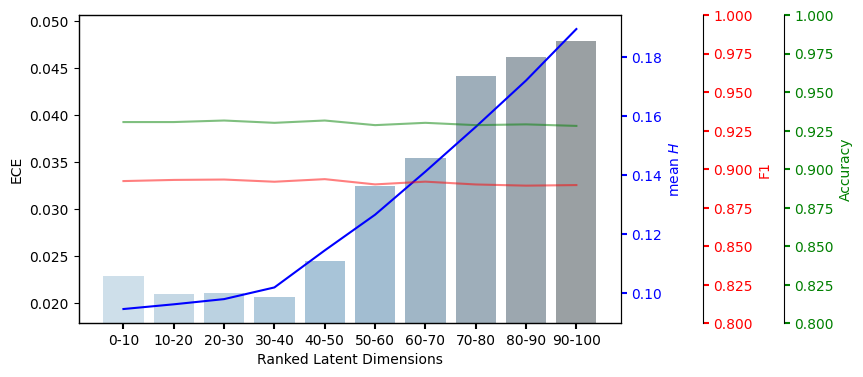}}
\subfloat[DistilBERT \texttt{CUE} on MultiNLI.]{
  \includegraphics[width=0.35\columnwidth]{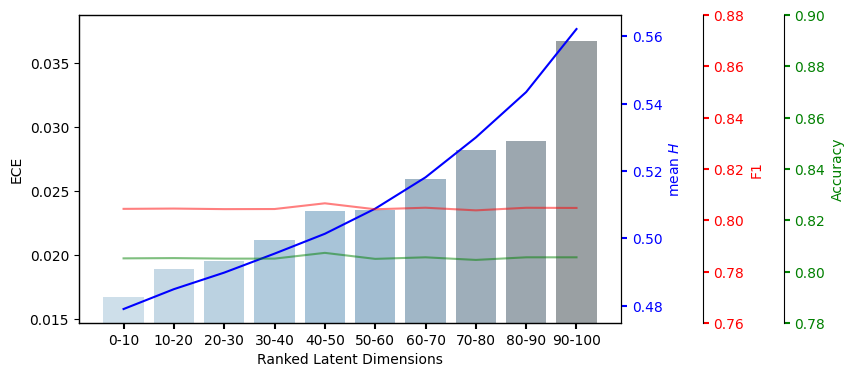}}\\
\subfloat[RoBERTa \texttt{CUE} on CoLA.]{
  \includegraphics[width=0.35\columnwidth]{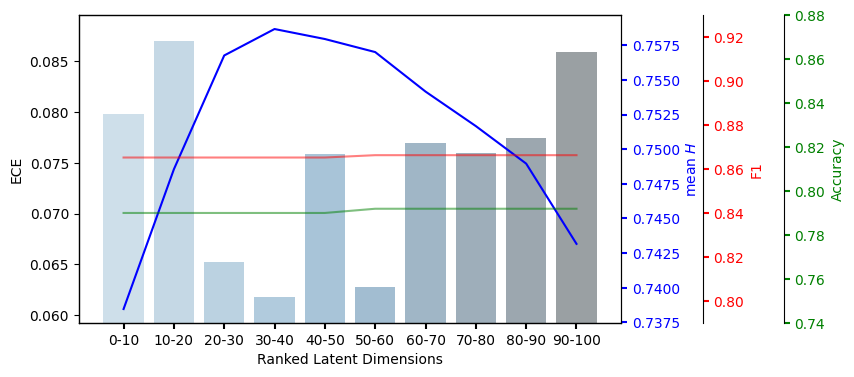}}
\subfloat[RoBERTa \texttt{CUE} on GoEmotions.]{
  \includegraphics[width=0.35\columnwidth]{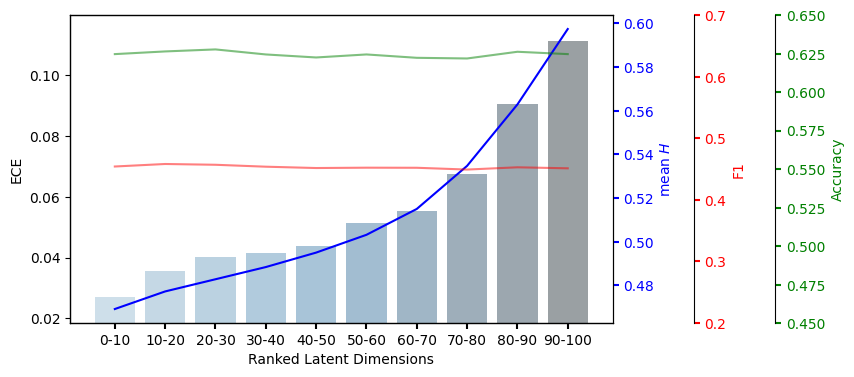}}\\
\subfloat[RoBERTa \texttt{CUE} on Emotion.]{
  \includegraphics[width=0.35\columnwidth]{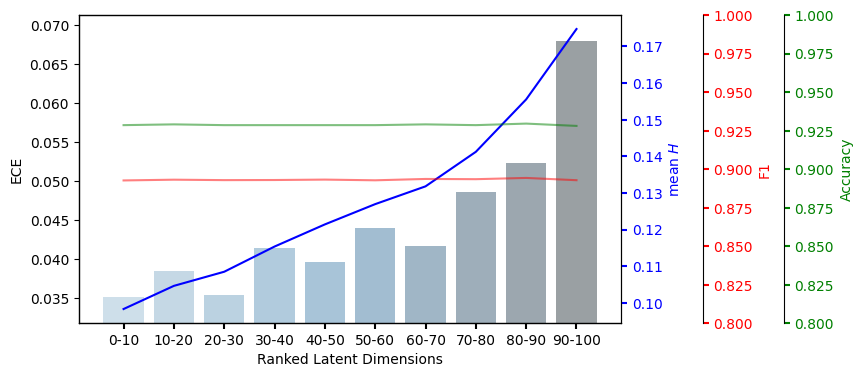}}
\subfloat[RoBERTa \texttt{CUE} on MultiNLI.]{
  \includegraphics[width=0.35\columnwidth]{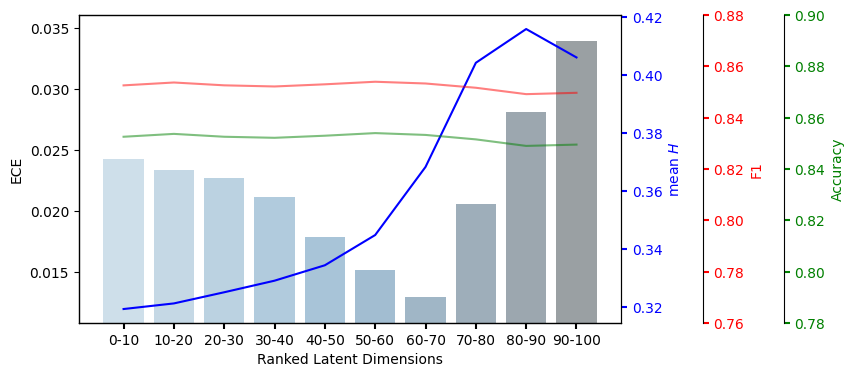}}
     \caption{Evaluation results by removing latent dimensions. The $x$-axis represents the index of \textbf{removed} dimensions ranked by their relevance to $\Delta \bm{e}_i$, smaller index number indicates the latent dimension is more similar. Histograms show the ECE scores after removing the corresponding latent dimensions. The blue curve shows the predictive entropy. The green and red curves show classification accuracy and F1, respectively.}
    \label{fig:latent_ablation}
\end{figure}

\subsection{Results with Latent Dimension Removal} 
As mentioned in \textsection{5.2}, we study the impact of removing ranked latent variable dimensions on the other three base models: ALBERT, DistilBERT and RoBERTa. As shown in Figure \ref{fig:latent_ablation}, we can observe the same trend of ECE score and average entropy increasing when removing latent dimensions ranked by their influential scores on almost all the models while keeping accuracy and macro F1 scores almost unchanged. This proves our \texttt{CUE} framework can be generalized to interpret the uncertainty via latent dimensions on various models and different datasets. On the CoLA dataset, both ALBERT and RoBERTa exhibit a similar pattern compared with the BERT model; we can also observe a peak for the average entropy.  
The graphs show our \texttt{CUE} can effectively distinguish the importance between latent dimensions, and thus we can use those dimensions to interpret token level uncertainty as discussed in \textsection{4.2}.

\subsection{Ablation Study}

As mentioned at the end of the \textsection{5}, we present an ablation study to investigate the contribution of various components in our framework.

\paragraph{Stability of training loss}
 
\begin{figure*}[h!]
  \centering
  \includegraphics[width=\columnwidth]{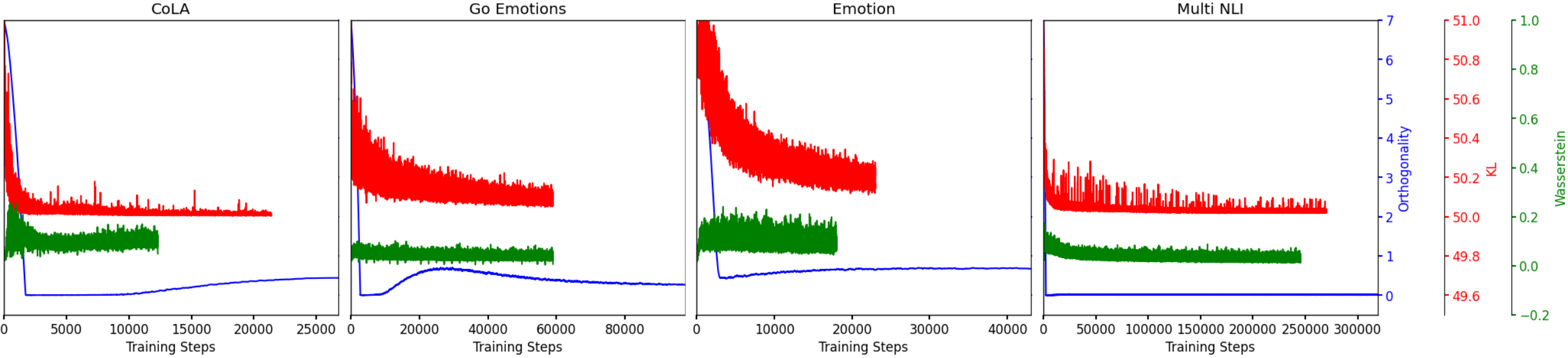}
\caption{Comparison of the stability of the total loss for three loss terms trained with BERT model on four datasets. Red: total loss with KL divergence loss; Green: total loss with Wasserstein loss; Blue: total loss with Orthogonality loss.}
\label{fig:loss_stability}
\end{figure*}

As discussed in \textsection{4.1}, our learning objective is implemented with four loss terms.
We investigate the training stability benefits from orthogonal regularisation by replacing the orthogonality loss with either a KL-divergence loss or a Wasserstein loss, where the KL-divergence loss encourages the distribution of latent variables to follow the prior standard Gaussian distribution and is widely used in general Variational Auto-encoders \citep{vae,scholar_vae}, while the Wasserstein loss enforces the latent variables to follow a Dirichlet distribution and is used in Wasserstein Auto Encoder (WAE) \citep{nan-etal-2019-topic, tolstikhin2018wasserstein}.
The total loss (including the reconstruction loss and the cross-entropy loss) curves during training are shown in Figure \ref{fig:loss_stability}. We observe that the total loss replaced by either the KL-divergence loss or the Wasserstein loss exhibits 
fluctuation during the training process across all datasets. 
On the contrary, the loss with orthogonal regularisation is very stable. We further show the evaluation results with various loss terms in Table \ref{table:loss} \footnote{Results reported are single run results, which we used to generate loss stability graphs.}. It can be observed that our proposed framework with the orthogonality loss achieves better ECE results compared to using KL or Wasserstein loss. 

\begin{table*}[h!]
\centering
\resizebox{\columnwidth}{!}{
\begin{tabular}{lllllllllllll}
\toprule
                 & \multicolumn{4}{c}{BERT \texttt{CUE} w/  Orthogonality}                                                              & \multicolumn{4}{c}{BERT \texttt{CUE} w/ KL}                                                              &\multicolumn{4}{c}{BERT \texttt{CUE} w/ Wasserstein }                          \\
\cmidrule(lr){2-5} \cmidrule(lr){6-9}   \cmidrule(lr){10-13}
Datasets & Acc    &  F1  & $\mathcal{H}$  & ECE$\downarrow$&Acc    &  F1  & $\mathcal{H}$  & ECE$\downarrow$&Acc    &  F1  & $\mathcal{H}$  & ECE$\downarrow$\\ \midrule
CoLA& 0.8130&0.7459&0.4986&\textbf{0.0640}&0.8072&0.7300&0.3407&0.1090&0.8044&0.7240&0.3499&0.1111\\
GoEmotions&0.6298&0.4661&0.4333&\textbf{0.0321}&0.6298&0.4752&0.3345&0.0695&0.6263&0.4608&0.3437&0.0600\\
Emotion&0.9255&0.8827&0.0984&\textbf{0.0322}&0.9270&0.8847&0.0518&0.0431&0.9275&0.8853&0.0510&0.0441\\
MultiNLI&0.8284&0.8278&0.3650&\textbf{0.0272}&0.8294&0.8290&0.3194&0.0418&0.8291&0.8286&0.3343&0.0344\\
\bottomrule
\end{tabular}}

\caption{Comparison of the performance of the BERT model fine-tuned with different loss terms on four datasets. }
\label{table:loss}
\end{table*}

\begin{figure*}[ht]
\centering
\subfloat[Additional KL divergence loss term.]{
    \includegraphics[width=0.4\columnwidth]{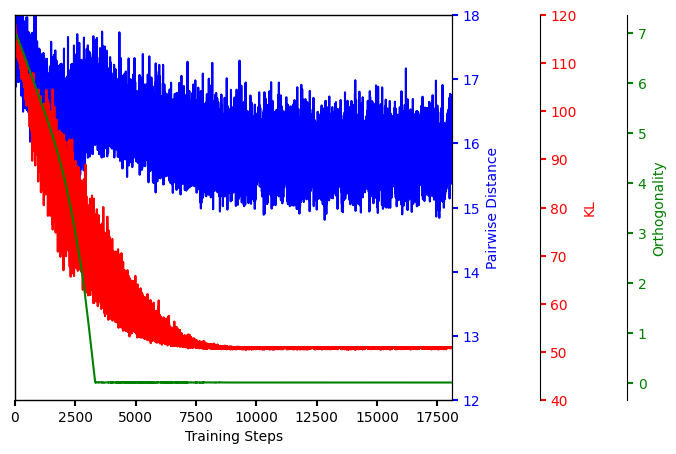}}
\subfloat[Additional Wasserstein loss term.]{
    \includegraphics[width=0.4\columnwidth]{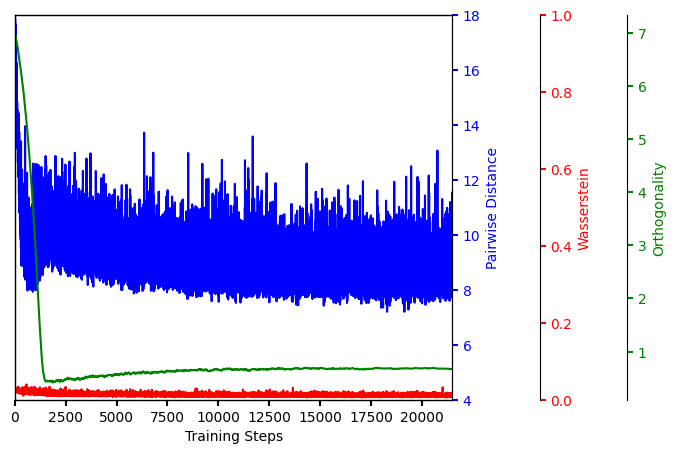}}
\caption{Comparison of BERT models trained on Emotions with additional loss term. Blue: Reconstruction loss; Red: KL loss in (a) and Wasserstein loss in (b); Green: Orthogonality loss.}
\label{fig:loss_compare_individual}
\end{figure*}

\paragraph{Training Loss Stability with Additional Loss Terms} 
We further examine the training loss stability when adding the KL or Wasserstein distance loss terms to our framework. We fine-tuned two BERT-base uncased \texttt{CUE} models on the Emotions dataset. It can be observed in Figure \ref{fig:loss_compare_individual} that the pairwise distance (i.e., the reconstruction loss) seems to be very unstable and keeps fluctuating during training while the orthogonality loss shows a stable decreasing trend and converges quickly. If we only compare the KL loss with the Wasserstein loss, we can see that the Wasserstein loss is more stable compared to KL. 
Our visualisation results show that the prior distributions assumed by the KL or the Wasserstein loss may not be suitable for reconstructing PLM-encoded representations, thus leading to higher ECE results compared to using the orthogonality constraint. 

\paragraph{Latent Space Orthogonality}
\begin{table*}[h!]
\centering
\resizebox{\columnwidth}{!}{
\begin{tabular}{lcccccccc}
\toprule
                 & \multicolumn{4}{c}{BERT \texttt{CUE} w/  Orthogonality}                                                              & \multicolumn{4}{c}{BERT \texttt{CUE} w/o Orthogonality}                             \\
\cmidrule(lr){2-5} \cmidrule(lr){6-9}   
Datasets &  Acc    &  F1  & $\mathcal{H}$  & ECE$\downarrow$    &  Acc    &  F1  & $\mathcal{H}$  & ECE$\downarrow$   \\ \midrule
CoLA      & 0.8123\small{±0.0012}&0.8762\small{±0.0007}&0.4991\small{±0.0032}&\textbf{0.0677}\small{±0.0056}& 0.8042\small{±0.0011}&0.7230\small{±0.0024}&0.3458\small{±0.0047}&0.1121\small{±0.0020} \\
GoEmotions& 0.6282\small{±0.0029}&0.4712\small{±0.0087}&0.4433\small{±0.0159}&\textbf{0.0326}\small{±0.0013}& 0.6291\small{±0.0019}&0.4652\small{±0.0037}&0.3432\small{±0.0014}&0.0615\small{±0.0023} \\
Emotion   & 0.9259\small{±0.0009}&0.8850\small{±0.0015}&0.1031\small{±0.0082}&\textbf{0.0289}\small{±0.0043}& 0.9268\small{±0.0003}&0.8848\small{±0.0006}&0.0519\small{±0.0009}&0.0430\small{±0.0016} \\
MultiNLI  & 0.8283\small{±0.0005}&0.8277\small{±0.0005}&0.3665\small{±0.0030}&\textbf{0.0262}\small{±0.0021}& 0.8281\small{±0.0009}&0.8277\small{±0.0009}&0.3317\small{±0.0009}&0.0370\small{±0.0010} \\
\bottomrule
\end{tabular}}
\caption{Comparison of results on BERT models trained with/without latent space orthogonality.}
\label{table:orthogonal}
\end{table*}
\begin{figure}[h!]
    \small
  \centering
  \includegraphics[width=0.6\columnwidth]{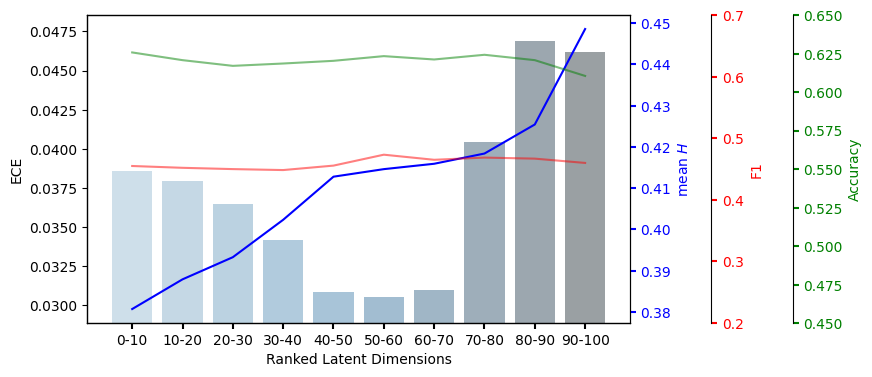}
  \caption{ECE and average entropy with latent dimension removal from the model trained without the orthogonality regulariser. The $x$-axis represents the index of \textbf{removed} dimensions ranked by their relevance to $\Delta \bm{e}_i$, smaller index number indicates the latent dimension is more similar. Histograms show the ECE scores after removing the corresponding latent dimensions. The blue curve shows the predictive entropy. The green and red curves show the classification accuracy and F1, respectively.}
  \label{fig:non_ort}
\end{figure} 

As explained in \textsection{4.1}, the orthogonality regulariser facilitates a better interpretation of the latent space. Shown in Table \ref{table:orthogonal}, we compare the overall performance between BERT models trained with and without latent space orthogonality. The PLMs fine-tuned with Eq. (8) outperform the counterparts without the orthogonality regularisation in ECE and average entropy on all four datasets. It is also interesting to find the f1 scores slightly decrease on the models trained without the orthogonality on almost all the datasets. Therefore, the orthogonality constraints ensure the decoder network to facilitate the same distribution on the latent space to generate reconstructed representations that lead to uncertain predictions. 


We performed a further ablation study to examine the interpretability of the latent space without being regularised by orthogonality. As shown in Figure \ref{fig:non_ort}, without the orthogonality loss term, there is no clear relationship between the tendency of ECE scores and the average entropy during the removal of latent dimensions ranked by their influential scores. Without orthogonality we are not able to maintain the distributional consistency from the latent representation space, hence we can see an obvious fluctuation in Accuracy and F1. Without a consistent tendency, it is thus 
difficult to investigate each latent dimension's importance and interpret the impact of each latent dimension on model predictive uncertainty. 

\bibliography{li_517-supp}